\documentclass{article}

\usepackage{microtype}
\usepackage{graphicx}
\usepackage{subfigure}
\usepackage{booktabs} % for professional tables

\usepackage{hyperref}

\usepackage[accepted]{icml2018}

\usepackage{float}
\usepackage{amssymb}
\usepackage{amsmath}
\usepackage{amsthm}
\usepackage{bbm}
\usepackage{csquotes}
\usepackage{color}
\usepackage{multirow}
\usepackage{array}
\usepackage{algorithm}
\usepackage{algorithmic}
\usepackage{caption}
\usepackage{nicefrac}
\usepackage[toc,title,page]{appendix}

\definecolor{mydarkblue}{rgb}{0,0.08,0.45}
\hypersetup{colorlinks=true,
    linkcolor=mydarkblue,
    citecolor=mydarkblue,
    filecolor=mydarkblue,
    urlcolor=mydarkblue}

\newcommand{\bm}[1]{\mathbf{#1}}
\newcommand{\expect}{\mathbb{E}}
\newcommand{\transpose}{\top}
\newcommand{\kldiv}{\mathrm{D}_{\mathrm{KL}}}
\newcommand{\klbars}{\,\|\,}
\DeclareMathOperator*{\tr}{tr}

\DeclareMathOperator*{\kvec}{vec}

\newcommand{\given}{\,|\,}
\newcommand{\normal}{\mathcal{N}}
\newcommand{\matrixNormal}{\mathcal{M} \mathcal{N}}
\newcommand{\zeroVec}{\mathbf{0}}
\newcommand{\ident}{\mathbf{I}}

\DeclareMathOperator*{\diag}{diag}

\newcommand{\data}{\mathcal{D}}
\newcommand{\elbo}{\mathcal{L}}
\newcommand{\params}{{\boldsymbol \theta}}
\newcommand{\initParams}{{\params_{0}}}

\newcommand{\lrate}{\alpha}
\newcommand{\lratePrec}{\beta}
\newcommand{\subspace}{\mathcal{S}}
\newcommand{\fisher}{\mathbf{F}}

\newcommand{\loss}{h}
\newcommand{\natgrad}{\tilde{\nabla}}
\newcommand{\cov}{{\boldsymbol{\Sigma}}}

\newcommand{\precision}{{\boldsymbol \Lambda}}
\newcommand{\mean}{{\boldsymbol \mu}}
\newcommand{\weights}{\mathbf{w}}
\newcommand{\qParams}{{\boldsymbol \phi}}
\newcommand{\klWeight}{\lambda}
\newcommand{\ndata}{N}
\newcommand{\target}{y}
\newcommand{\inputVec}{\mathbf{x}}
\newcommand{\priorVar}{\eta}
\newcommand{\lrateMod}{\tilde{\lrate}}
\newcommand{\lratePrecMod}{\tilde{\lratePrec}}
\newcommand{\fisherAverage}{\bar{\fisher}}
\newcommand{\derivative}{\mathcal{D}}
\newcommand{\grad}{\mathbf{g}_l}

\newcommand{\fisherVec}{\mathbf{f}}
\newcommand{\fisherVecAverage}{\bar{\fisherVec}}

\DeclareMathOperator*{\Cov}{Cov}

\icmltitlerunning{Noisy Natural Gradient as Variational Inference}

\begin{document}

\twocolumn[
\icmltitle{Noisy Natural Gradient as Variational Inference}

% It is OKAY to include author information, even for blind
% submissions: the style file will automatically remove it for you
% unless you've provided the [accepted] option to the icml2018
% package.

% List of affiliations: The first argument should be a (short)
% identifier you will use later to specify author affiliations
% Academic affiliations should list Department, University, City, Region, Country
% Industry affiliations should list Company, City, Region, Country

% You can specify symbols, otherwise they are numbered in order.
% Ideally, you should not use this facility. Affiliations will be numbered
% in order of appearance and this is the preferred way.
\icmlsetsymbol{equal}{*}

\begin{icmlauthorlist}
\icmlauthor{Guodong Zhang}{equal,uoft,vector}
\icmlauthor{Shengyang Sun}{equal,uoft,vector}
\icmlauthor{David Duvenaud}{uoft,vector}
\icmlauthor{Roger Grosse}{uoft,vector}
\end{icmlauthorlist}

\icmlaffiliation{uoft}{University of Toronto}
\icmlaffiliation{vector}{Vector Institute}

\icmlcorrespondingauthor{Guodong Zhang}{gdzhang@cs.toronto.edu}
\icmlcorrespondingauthor{Shengyang Sun}{ssy@cs.toronto.edu}

% You may provide any keywords that you
% find helpful for describing your paper; these are used to populate
% the "keywords" metadata in the PDF but will not be shown in the document
\icmlkeywords{Bayesian Neural Networks, Natural Gradient, Machine Learning, ICML}

\vskip 0.3in
]

% this must go after the closing bracket ] following \twocolumn[ ...

% This command actually creates the footnote in the first column
% listing the affiliations and the copyright notice.
% The command takes one argument, which is text to display at the start of the footnote.
% The \icmlEqualContribution command is standard text for equal contribution.
% Remove it (just {}) if you do not need this facility.

%\printAffiliationsAndNotice{}  % leave blank if no need to mention equal contribution
\printAffiliationsAndNotice{\icmlEqualContribution} % otherwise use the standard text.

\begin{abstract}
Variational Bayesian neural nets combine the flexibility of deep learning with Bayesian uncertainty estimation.
Unfortunately, there is a tradeoff between cheap but simple variational families (e.g.~fully factorized) or expensive and complicated inference procedures.
We show that natural gradient ascent with adaptive weight noise implicitly fits a variational posterior to maximize the evidence lower bound (ELBO).
This insight allows us to train full-covariance, fully factorized, or matrix-variate Gaussian variational posteriors using noisy versions of natural gradient, Adam, and K-FAC, respectively, making it possible to scale up to modern ConvNets.
On standard regression benchmarks, our noisy K-FAC algorithm makes better predictions and matches Hamiltonian Monte Carlo's predictive variances better than existing methods.
Its improved uncertainty estimates lead to more efficient exploration in active learning, and intrinsic motivation for reinforcement learning.
\end{abstract}

\section{Introduction}

Combining deep learning with Bayesian uncertainty estimation has the potential to fit flexible and scalable models that are resistant to overfitting~\citep{mackay1992practical,neal1995bayesian,hinton1993keeping}.
Stochastic variational inference is especially appealing because it closely resembles ordinary backprop~\citep{graves2011practical,blundell2015weight}, but such methods typically impose restrictive factorization assumptions on the approximate posterior, such as fully independent weights.
There have been attempts to fit more expressive approximating distributions which capture correlations such as matrix-variate Gaussians~\citep{louizos2016structured,sun2017learning} or multiplicative normalizing flows~\citep{louizos2017multiplicative}, but fitting such models can be expensive without further approximations.

In this work, we introduce and exploit a surprising connection between natural gradient descent~\citep{amari1998natural} and variational inference.
In particular, several approximate natural gradient optimizers have been proposed which fit tractable approximations to the Fisher matrix to gradients sampled during training~\citep{kingma2014adam, martens2015optimizing}.
While these procedures were described as natural gradient descent on the weights using an \emph{approximate} Fisher matrix, we reinterpret these algorithms as natural gradient on a variational posterior using the \emph{exact} Fisher matrix.
Both the weight updates and the Fisher matrix estimation can be seen as natural gradient ascent on a unified evidence lower bound (ELBO), analogously to how Neal and Hinton~\citep{neal1998view} interpreted the E and M steps of Expectation-Maximization (E-M) as coordinate ascent on a single objective.

Using this insight, we give an alternative training method for variational Bayesian neural networks.
For a factorial Gaussian posterior, it corresponds to a diagonal natural gradient method with weight noise, and matches the performance of Bayes By Backprop~\citep{blundell2015weight}, but converges faster.
We also present noisy K-FAC, an efficient and GPU-friendly method for fitting a full matrix-variate Gaussian posterior, using a variant of Kronecker-Factored Approximate Curvature (K-FAC)~\citep{martens2015optimizing} with correlated weight noise.

\begin{figure*}[t]
\centering
\hspace{-1.5em}%
\subfigure[Fully factorized] { \label{fig:precision_bbb} 
\includegraphics[height=2.8cm]{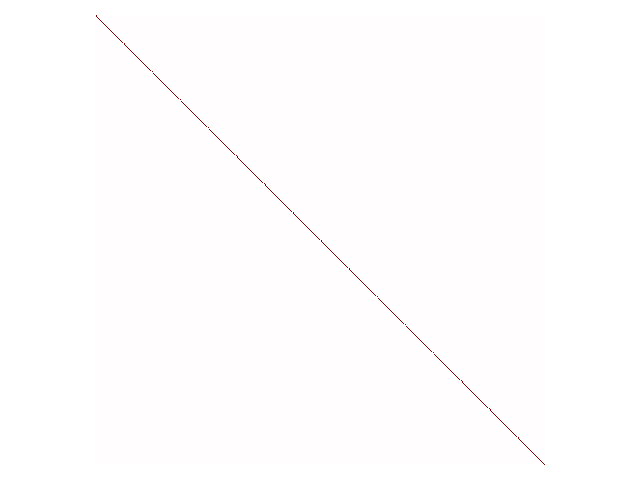} 
} 
\hspace{-1.5em}%
\subfigure[Matrix-variate] { \label{fig:precision_mvg} 
\includegraphics[height=2.8cm]{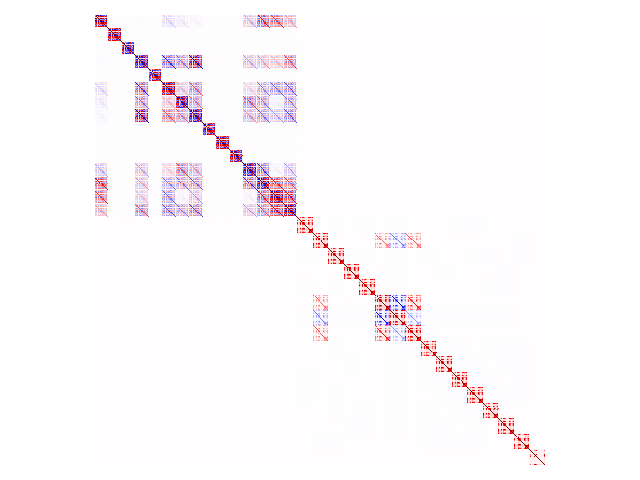} 
}
\hspace{-1.5em}%
\subfigure[Block tridiagonal] { \label{fig:precision_tri} 
\includegraphics[height=2.8cm]{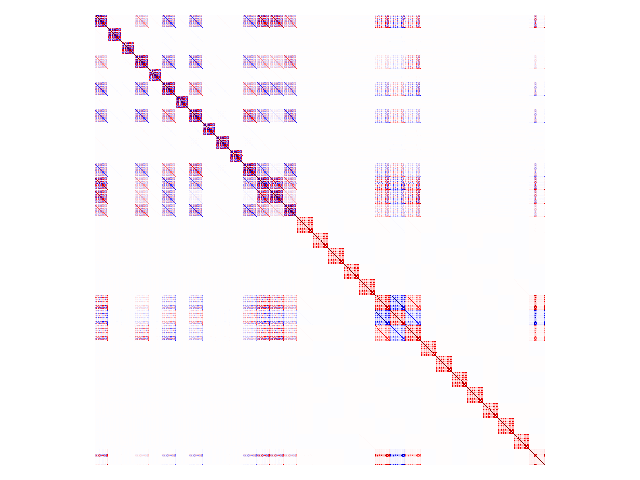} 
} 
\hspace{-1.5em}%
\subfigure[Full covariance] { \label{fig:precision_full} 
\includegraphics[height=2.8cm]{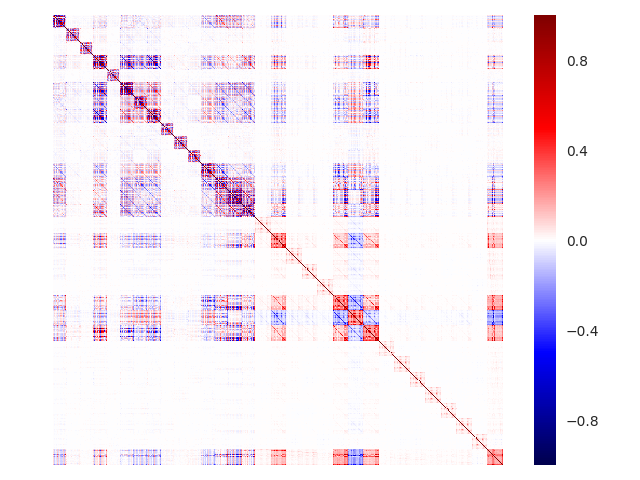} 
} 
\vspace{-1em}
\caption{ Normalized precision matrices for Gaussian variational posteriors trained using noisy natural gradient. We used a network with 2 hidden layers of 15 units each, trained on the Boston housing dataset.} 
\label{fig:precision} 
\end{figure*}

\section{Background}

\subsection{Variational Inference for Bayesian Neural Networks} 
Given a dataset $\data = \{(\inputVec_i,\target_i)_{i=1}^n\}$,
a Bayesian neural net (BNN) is defined in terms of a prior $p(\weights)$ on the weights, as well as the likelihood $p(\data \given \weights)$.
% Performing inference on the BNN requires integrating over the intractable posterior distribution $p(\weights\given\data)$.
Variational Bayesian methods~\citep{hinton1993keeping,graves2011practical,blundell2015weight} attempt to fit an approximate posterior $q(\weights)$ to maximize the evidence lower bound (ELBO):
\begin{equation}
\label{eq:elbo}
\elbo[q] = \expect_{q}[\log p(\data \given \weights)] - \klWeight \kldiv(q(\weights)\klbars p(\weights)) %\\ & = \expect[\log p(\data \given \weights)] + \klWeight \expect[\log p(\weights)] + \klWeight \entropy(q)
\end{equation}
where $\klWeight$ is a regularization parameter and $\qParams$ are the parameters of the variational posterior.
Proper Bayesian inference corresponds to $\klWeight=1$, but other values may work better in practice on some problems.
%The most commonly used variational BNN training method is Bayes By Backprop (BBB)~\citep{blundell2015weight}, which uses a fully factorized Gaussian approximation to the posterior, i.e.~$q(\weights) = \normal(\weights;\mean, \diag(\stdVec^2))$.
%The variational parameters $\qParams = (\mean, \stdVec^2)$ are adapted using stochastic gradients of $\elbo$ obtained using the reparameterization trick~\citep{kingma2013auto}.
%The resulting updates for $\mean$ look like ordinary backpropagation updates for the weights of a non-Bayesian neural network, except that $\weights$ is sampled from the variational posterior.
%Another interpretation of BBB regards $\mean$ as a point estimate of the weights and $\stdVec$ as the standard deviations of Gaussian noise added independently for each training example.
%This similarity to ordinary neural net training is a big part of the appeal of BBB and other variational BNN approaches.

\subsection{Gradient Estimators for Gaussian Distribution}
To optimize the ELBO, we must estimate the derivative of eq.~\eqref{eq:elbo} w.r.t.\ variational parameters $\qParams$.
%This is a special case of the problem of estimating $\frac{\partial }{\partial \qParams} \expect_{q_{\qParams}(\weights)}[f(\weights)]$ for some $f$.
%
%There are two standard approaches to estimating gradients of expectations of functions of random variables.
%The most generally-applicable method is the score-function estimator, or REINFORCE~\cite{williams1992simple}.
%This method requires only evaluation of $f(\weights)$ and not its derivatives, so can used even if $f(\weights)$ is discontinuous. 
%
The standard approach uses the pathwise derivative estimator, also known as the reparameterization trick~\citep{williams1992simple, blundell2015weight, kingma2013auto, rezende2014stochastic}.
%This estimator is especially useful when the parameters $\qParams$ are high dimensional.
%It usually has lower variance than REINFORCE, while remaining unbiased.
%However, it requires the derivative $\nicefrac{\partial f}{\partial \weights}$.
%
However, in the case of Gaussian distribution with parameters $\qParams= \{\mean, \cov\}$, there is another estimator given by \citet{opper2009variational}:
\begin{equation}
\begin{aligned}
\nabla_{\mean} \expect_{\normal(\mean, \cov)}[f(\weights)] &= \expect_{\normal(\mean, \cov)}\left[\nabla_{\weights} f(\weights)\right] \\
\nabla_{\cov} \expect_{\normal(\mean, \cov)}[f(\weights)] &= \expect_{\normal(\mean, \cov)}\left[\nabla_{\weights}^2 f(\weights)\right]
\label{eq:grad-estimator}
\end{aligned}
\end{equation}
which are due to \citet{bonnet1964transformations} and \citet{price1958useful}, respectively.
Both equations can be proved through integration by parts.
In the case of Gaussian distribution, eq.~\eqref{eq:grad-estimator} is equivalent to the pathwise derivative estimator for $\mean$. 
\subsection{Natural Gradient} 
\label{sec:natgrad}

Natural gradient descent is a second-order optimization method originally proposed by \citet{amari1997neural}.
There are two variants of natural gradient commonly used in machine learning, which do not have standard names, but which we refer to as natural gradient for point estimation (NGPE) and natural gradient for variational inference (NGVI).
%While both methods are broadly applicable, we limit the present discussion to neural networks for simplicity.

In natural gradient for point estimation (NGPE), we assume the neural network computes a predictive distribution $p(\target|\inputVec;\weights)$ and we wish to maximize a cost function $\loss(\weights)$, which may be the data log-likelihood.
The natural gradient is the direction of steepest ascent in the Fisher information norm, and is given by $\natgrad_\weights \loss = \fisher^{-1} \nabla_\weights \loss$, where
$\fisher = \Cov_{\inputVec \sim p_{\data}, \target \sim p(\target|\inputVec, \weights)} \left[ \nabla_\weights \log p(\target|\inputVec, \weights) \right]$,
and the covariance is with respect to $\inputVec$ sampled from the data distribution and $\target$ sampled from the model's predictions.
NGPE is typically justified as a way to speed up optimization; 
%the details are inessential to this paper; 
see \citet{martens2014new} for a comprehensive overview.
%Because the dimension of $\fisher$ is the number of parameters, which can be in the tens of millions for modern neural networks, exact computation of the natural gradient is typically infeasible, and approximations are required (see below). 

We now describe natural gradient for variational inference (NGVI) in the context of BNNs.
We wish to fit the parameters of a variational posterior $q(\weights)$ to maximize the ELBO (eq.~\eqref{eq:elbo}).
Analogously to the point estimation setting, the natural gradient is defined as $\natgrad_\qParams \elbo = \fisher^{-1} \nabla_\qParams \elbo$; but in this case, $\fisher$ is the Fisher matrix of $q$, i.e. $\fisher = \Cov_{\weights \sim q}\left[ \nabla_\qParams \log q(\weights ; \qParams) \right]$.
Note that in contrast with point estimation, $\fisher$ is a metric on $\qParams$, rather than $\weights$, and its definition doesn't directly involve the data.
Interestingly, because $q$ is chosen to be tractable, the natural gradient can be computed exactly, and in many cases is even simpler than the ordinary gradient.
%As an important example, \citet{hoffman2013stochastic} used natural gradient to scale up variational Bayes for latent variable models such as LDA, and found that the natural gradient ascent updates closely resemble the EM-like updates used in variational Bayes.

In general, NGPE and NGVI need not behave similarly; however, in Section~\ref{sec:method}, we show that in the case of Gaussian variational posteriors, the two are closely related.

\subsection{Kronecker-Factored Approximate Curvature} 

As modern neural networks may contain millions of parameters, computing and storing the exact Fisher matrix and its inverse is impractical.
Kronecker-factored approximate curvature (K-FAC)~\citep{martens2015optimizing} uses a Kronecker-factored approximation to the Fisher to perform efficient approximate natural gradient updates.
Considering $l$th layer in the neural network whose input activations are $\bm{a}_l \in \mathbb{R}^{n_1}$, weight $\bm{W}_l \in \mathbb{R}^{n_1 \times n_2}$, and output $\bm{s}_l \in \mathbb{R}^{n_2}$, we have $\bm{s}_l = \bm{W}_l^T\bm{a}_l$. For simplicity, we define the following additional notation:
\begin{equation*}
\derivative v = \nabla_{v} \log p(\target | \inputVec, \weights) \text{ and } \grad = \derivative \bm{s}_l
\end{equation*}
Therefore, the weight gradient is $\derivative{\bm{W}_l} = \bm{a}_l\grad^T$. With this gradient formula, K-FAC decouples this layer's fisher matrix $\fisher_l$ by approximating $\bm{a}_l$ and $\grad$ as independent:
\begin{equation}
\label{eq:kfac}
\begin{aligned}
	\fisher_l &= \expect[\mathrm{vec}\{\derivative{\bm{W}_l} \}\mathrm{vec}\{\derivative{\bm{W}_l} \}^{\transpose}] 
	= \expect[\grad\grad^{\transpose} \otimes \bm{a}_l\bm{a}_l^{\transpose}] \\
    &\approx \mathbb{E} [\grad\grad^{\transpose}] \otimes \mathbb{E}[\bm{a}_l\bm{a}_l^{\transpose}] = \bm{S}_l \otimes \bm{A}_l = \tilde{\fisher}_l
\end{aligned}
\end{equation}
%Where $\bm{A}_l = \mathbb{E}[\bm{a}\bm{a}^T]$ and $\bm{S}_l = \mathbb{E} [\{\nabla_\bm{s} \loss \}\{\nabla_\bm{s} \loss \}^T]$.
Further, assuming between-layer independence, the whole fisher matrix $\tilde{\fisher}$ can be approximated as block diagonal consisting of layerwise fisher matrices $\tilde{\fisher}_l$. Decoupling $\tilde{\fisher}_l$ into $\bm{A}_l$ and $\bm{S}_l$ not only avoids the quadratic storage cost of the exact Fisher, but also enables tractable computation of the approximate natural gradient:
\begin{equation}
\label{eq:ng-kfac}
\begin{aligned}
	\tilde{\fisher}_{l}^{-1}\mathrm{vec}\{\nabla_{\bm{W}_l}\loss\} &= \bm{S}_l^{-1} \otimes \bm{A}_l^{-1} \mathrm{vec}\{\nabla_{\bm{W}_l}\loss\} \\ 
	&= \mathrm{vec}[\bm{A}_l^{-1} \nabla_{\bm{W}_l}\loss \bm{S}_l^{-1}]
\end{aligned}
\end{equation}
As shown by eq.~\eqref{eq:ng-kfac}, computing natural gradient using K-FAC only consists of matrix transformations comparable to size of $\bm{W}_l$, making it very efficient.

\section{Variational Inference using Noisy Natural Gradient}
\label{sec:method}

In this section, we draw a surprising relationship between natural gradient for point estimation (NGPE) of the weights of a neural net, and natural gradient for variational inference (NGVI) of a Gaussian posterior.
(These terms are explained in Section~\ref{sec:natgrad}.)
In particular, we show that the NGVI updates can be approximated with a variant of NGPE with adaptive weight noise which we term \emph{Noisy Natural Gradient (NNG)}.
This insight allows us to train variational posteriors with a variety of structures using noisy versions of existing optimization algorithms (see Figure~\ref{fig:precision}).

In NGVI, our goal is to maximize the ELBO $\elbo$ (eq.~\eqref{eq:elbo}) with respect to the parameters $\qParams$ of a variational posterior distribution $q(\weights)$.
We assume $q$ is a multivariate Gaussian parameterized by $\qParams=(\mean, \cov)$.
Building on eq.~\eqref{eq:grad-estimator}, we determine the natural gradient of the ELBO with respect to $\mean$ and the precision matrix $\precision = \cov^{-1}$ (see supplement for details):
\begin{align}
\natgrad_\mean \elbo &= \precision^{-1} \expect_q \left[ \nabla_\weights \log p(\data \given \weights) + \klWeight \nabla_\weights \log p(\weights) \right] \label{eq:ng-update} \\
\natgrad_\precision \elbo &= -\expect_q \left[ \nabla^2_\weights \log p(\data \given \weights) + \klWeight \nabla^2_\weights \log p(\weights) \right] - \klWeight \precision \notag
\end{align}
%
%Here, $\klWeight$ is the weighting of the KL term in eq.~\eqref{eq:elbo}.
We make several observations.
First, the term inside the expectation in eq.~\eqref{eq:ng-update} is the gradient for MAP estimation of $\weights$.
Second, the update for $\mean$ is preconditioned by $\precision^{-1}$, which encourages faster movement in directions of higher posterior uncertainty.
Finally, the fixed point equation for $\precision$ is given by
\begin{equation}
\precision = -\expect_q \left[ \frac{1}{\lambda} \nabla^2_\weights \log p(\data \given \weights) + \nabla^2_\weights \log p(\weights) \right]
\end{equation}
Hence, if $\klWeight = 1$, $\precision$ will tend towards the expected Hessian of $-\log p(\weights, \data)$, so the update rule for $\mean$ will somewhat resemble a Newton-Raphson update. For simplicity, we further assume a spherical Gaussian prior $\weights \sim \normal(\zeroVec, \priorVar \ident)$, then $\nabla^2_\weights \log p(\weights) = -\priorVar^{-1} \ident$.
%Smaller values of $\klWeight$ result in increased weighting of the log-likelihood Hessian, and hence a more concentrated posterior.
%\citet{khan2017variational} independently derived a similar stochastic Newton update; see Section~\ref{sec:related}.
%Based on these formulas, we derive the following stochastic natural gradient ascent updates based on 
In each iteration, we sample $(\inputVec, \target) \sim p_{\data}$ and $\weights \sim q$ and apply a stochastic natural gradient update based on eq.~\eqref{eq:ng-update}:
\begin{align}
\label{eq:ng_svi}
\mean &\gets \mean + \lrate \precision^{-1} \left[ \derivative{\weights} - \frac{\klWeight}{\ndata \priorVar} \weights \right] \\
\precision &\gets \left(1 - \frac{\klWeight \lratePrec}{\ndata} \right) \precision - \lratePrec \left[ \nabla_{\weights}^2\log p(\target|\inputVec, \weights) - \frac{\klWeight}{\ndata\priorVar} \ident \right] \notag
\end{align}
where $\lrate$ and $\lratePrec$ are separate learning rates for $\mean$ and $\precision$, and $\ndata$ is the number of training examples. Roughly speaking, the update rule for $\precision$ corresponds to an exponential moving average of the Hessian, and the update rule for $\mean$ is a stochastic Newton step using $\precision$. 

This update rule has two problems.
First, the log-likelihood Hessian may be hard to compute,  and is undefined at some points for neural nets which use not-everywhere-differentiable activation functions such as ReLU.
Second, if the negative log-likelihood is non-convex (as is the case for neural networks), the Hessian could have negative eigenvalues. 
%This update rule has the problems that the log-likelihood Hessian can be hard to compute, and that the resulting $\precision$ might not be PSD.
so the update may result in $\precision$ which is not positive semidefinite.
We circumvent both of these problems by approximating the negative log-likelihood Hessian with the NGPE Fisher matrix $\fisher = \Cov_{\inputVec \sim p_{\data}, \target \sim p(\target|\inputVec,\weights)}(\derivative \weights)$:
\begin{equation}
\precision \gets \left(1 - \frac{\klWeight \lratePrec}{\ndata} \right)\precision 
+ \lratePrec \left[ \derivative{\weights} {\derivative{\weights}}^\transpose +\frac{\klWeight}{\ndata \priorVar} \ident \right]	
\label{eq:graves-update}
\end{equation}
This approximation guarantees that $\precision$ is positive semidefinite, and it allows for tractable approximations such as K-FAC (see below).
In the context of BNNs, approximating the log-likelihood Hessian with the Fisher was first proposed by~\citet{graves2011practical}, so we refer to it as the \emph{Graves approximation}.
In the case where the output layer of the network represents the natural parameters of an exponential family distribution (as is typical in regression or classification), the Graves approximation can be justified in terms of the generalized Gauss-Newton approximation to the Hessian; see \citet{martens2014new} for details.\footnote{eq.~\eqref{eq:graves-update} leaves ambiguous what distribution the gradients are sampled from. Throughout our experiments, we sample the targets from the model's predictions, as done in K-FAC \citep{martens2015optimizing}. The resulting $\fisher$ is known as the \emph{true Fisher}. The alternative is to use the SGD gradients, giving the \emph{empirical Fisher}. The true Fisher is a better approximation to the Hessian~\citep{martens2014new}.}

\subsection{Simplifying the Update Rules}
We have now derived a stochastic natural gradient update rule for Gaussian variational posteriors.
In this section, we rewrite the update rules in order to disentangle hyperparameters and highlight relationships with NGPE.
First, if the prior variance $\priorVar$ is fixed\footnote{For simplicity, we assume the prior is a spherical Gaussian and its variance $\priorVar$ is fixed. Otherwise, we can keep an exponential moving average of the prior Hessian.}, then $\precision$ is a damped version of the moving average of the Fisher matrix and we can rewrite the update eq.~\eqref{eq:graves-update}:
\begin{equation}
\begin{aligned}
\precision & = \frac{\ndata}{\klWeight} \fisherAverage + \priorVar^{-1}\ident \\
\fisherAverage & \gets (1 - \lratePrecMod) \fisherAverage + \lratePrecMod \derivative\weights {\derivative\weights}^\transpose \\
%\priorHessianAverage & \gets (1 - \lratePrecMod ) \priorHessianAverage - \lratePrecMod \priorHessian
\end{aligned}
\label{eq:ng-svi-G}
\end{equation}
% 
%For a spherical Gaussian prior $\weights \sim \normal(\zeroVec, \priorVar \ident)$, we have $\priorHessian = \priorVar^{-1} \ident$.

In eq.~\eqref{eq:ng-svi-G}, we avoid an awkward interaction between the KL weight $\klWeight$ and the learning rates $\lrate, \lratePrec$ by writing the update rules in terms of alternative learning rates $\lrateMod = \lrate \klWeight / \ndata$ and $\lratePrecMod = \lratePrec \klWeight / \ndata$.
We also rewrite the update rule for $\mean$:
\begin{equation}
\mean \gets \mean + \lrateMod \left(\fisherAverage + \frac{\klWeight}{\ndata \priorVar} \ident \right)^{-1} \left[ \derivative{\weights} - \frac{\klWeight}{\ndata \priorVar} \weights \right]
\end{equation}
Observe that if $\mean$ is viewed as a point estimate of the weights, this update rule resembles NGPE with an exponential moving average of the Fisher matrix.
The differences are that the Fisher matrix $\fisher$ is damped by adding $\frac{\klWeight}{\ndata \priorVar} \ident$, and that the weights are sampled from $q$, which is a Gaussian with covariance $\cov = (\frac{\ndata}{\klWeight} \fisherAverage + \priorVar^{-1}\ident)^{-1}$.
Because our update rule so closely resembles NGPE with correlated weight noise, we refer to this method as Noisy Natural Gradient (NNG).

\subsection{Damping}

%In the special case of a spherical Gaussian prior\footnote{If the prior variance $\priorVar$ is not fixed, we need to keep an exponential moving average of it.}, we have that $p(\weights) = \normal(\weights ; \zeroVec, \priorVar \ident)$, and therefore $\nabla^2_\weights \log p(\weights) = -\priorVar^{-1} \ident$. 
%Therefore, $\priorHessianAverage = \bar{\priorVar}^{-1} \ident$, where $\bar{\priorVar}^{-1}$ is an exponential moving average of the prior precision.
%(If $\priorVar$ is fixed, then we have simply $\priorHessianAverage = \priorVar^{-1} \ident$).
Interestingly, in second-order optimization, it is very common to dampen the updates by adding a multiple of the identity matrix to the curvature before inversion in order to compensate for errors in the quadratic approximation to the cost.
NNG automatically achieves this effect, with the strength of the damping being $\klWeight / \ndata \priorVar$; we refer to this as \emph{intrinsic damping}.
In practice, it may be advantageous to add additional \emph{extrinsic damping} for purposes of stability.
%However, we found we did not need to do this in regression experiments.\footnote{We speculate that because the precision $\precision$ is fit using variational inference rather than a Taylor approximation, it is encouraged to reflect the global shape of the local mode of the distribution, helping to stabilize the update. \GD{And in regression, the number of training data is small which leads to strong intrinsic damping.}}

%The derivation above demonstrates the connection between the covariance of the variational distribution and the Fisher information.
%Imposing different structures on Fisher corresponds to varying posterior distributions.
%As a full covariance multivariate Gaussian posterior is computationally impractical for all but the smallest networks, next we will show several simplified Fisher structures and connect them to different posterior families.

\subsection{Fitting Fully Factorized Gaussian Posteriors with Noisy Adam}
\label{sec:ffg}
The discussion so far has concerned NGVI updates for a full covariance Gaussian posterior.
Unfortunately, the number of parameters needed to represent a full covariance Gaussian is of order $(\dim \weights)^2$.
Since it can be in the millions even for a relatively small network, representing a full covariance Gaussian is impractical.
There has been much work on tractable approximations to second-order optimization. In the context of NNG, imposing structure on $\fisher$ also imposes structure on the form of the variational posterior. We now discuss two kinds of structure one can impose.

Perhaps the simplest approach is to approximate $\fisher$ with a diagonal matrix $\mathrm{diag}(\fisherVec)$, as done by Adagrad~\citep{duchi2011adaptive} and Adam~\citep{kingma2014adam}.
For our NNG approach, this yields the following updates:
\begin{equation}
\begin{aligned}
	\mean &\gets \mean + \lrateMod \left[\derivative{\weights} - \frac{\klWeight}{\ndata \priorVar} \weights \right] / \left( \fisherVecAverage + \frac{\klWeight}{\ndata \priorVar} \right) \\
	\fisherVecAverage &\gets (1 - \lratePrecMod) \fisherVecAverage + \lratePrecMod {\derivative{\weights}}^2
	\label{eq:ng-mf}
\end{aligned}	
\end{equation}
These update rules are similar in spirit to methods such as Adam, but with the addition of adaptive weight noise.
We note that these update rules also differ from Adam in some details: (1) Adam keeps exponential moving averages of the gradients, which is equivalent to momentum, and (2) Adam applies the square root to the entries of $\fisherVec$ in the denominator.
We define \emph{noisy Adam} by adding momentum term to be consistent with Adam. We regard difference (2) inessential. The choice of squaring or divison may affect optimization performance, but they don't change the fixed points, i.e.~they are fitting the same functional form of the variational posterior using the same variational objective. The full procedure is given in Alg.~\ref{alg:noisy-adam}.
%We note that these modifications may affect optimization performance, but they don't change the fixed points, i.e.~they are fitting the same functional form of the variational posterior using the same variational objective. 

\subsection{Fitting Matrix Variate Gaussian Posteriors with Noisy K-FAC}
There has been much interest in fitting BNNs with matrix-variate Gaussian (MVG) posteriors\footnote{When we refer to a BNN with an ``MVG posterior'', we mean that the weights in different layers are independent, and the weights for each layer follow an MVG distribution.} in order to compactly capture posterior correlations between different weights~\citep{louizos2016structured,sun2017learning}.
Let $\mathbf{W}_l$ denote the weights for one layer of a fully connected network.
An MVG distribution is a Gaussian distribution whose covariance is a Kronecker product, i.e.~$\matrixNormal(\mathbf{W} ; \mathbf{M}, \cov_1, \cov_2) = \normal(\kvec(\mathbf{W}) ; \kvec(\mathbf{M}), \cov_2 \otimes \cov_1)$. 
%One can sample from an MVG distribution by sampling a matrix $\mathbf{E}$ of i.i.d.~standard Gaussians and then taking $\mathbf{W} = \mathbf{M} + \cov_1^{1/2} \mathbf{E} \cov_2^{1/2}$, where $\mean = \kvec(\mathbf{M})$.
MVGs are potentially powerful due to their compact representation\footnote{If $\mathbf{W}$ is of size $m \times n$, then the MVG covariance requires approximately $m^2/2 + n^2/2$ parameters to represent, in contrast with a full covariance matrix over $\weights$, which would require $m^2n^2/2$.} of posterior covariances between weights.
However, fitting MVG posteriors is difficult, since computing the gradients and enforcing the positive semidefinite constraint for $\cov_1$ and $\cov_2$ typically requires expensive matrix operations such as inversion.
Therefore, existing methods for fitting MVG posteriors typically impose additional structure such as diagonal covariance~\citep{louizos2016structured} or products of Householder transformations~\citep{sun2017learning} to ensure efficient updates.

We observe that K-FAC~\citep{martens2015optimizing} uses a Kronecker-factored approximation to the Fisher matrix for each layer's weights, as in eq.~\eqref{eq:kfac}. By plugging this approximation in to eq.~\eqref{eq:ng-svi-G}, we obtain an MVG posterior.
In more detail, each block obeys the Kronecker factorization $\mathbf{S}_l \otimes \mathbf{A}_l$, where $\mathbf{A}_l$ and $\mathbf{S}_l$ are the covariance matrices of the activations and pre-activation gradients, respectively.
K-FAC estimates $\mathbf{A}_l$ and $\mathbf{S}_l$ online using exponential moving averages which, conveniently for our purposes, are closely analogous to the exponential moving averages defining $\fisherAverage$ in eq.~\eqref{eq:ng-svi-G}:
\begin{equation}
\label{eq:running-average}
\begin{aligned}
	\bar{\mathbf{A}}_l &\gets (1-\lratePrecMod) \bar{\mathbf{A}}_l + \lratePrecMod \bm{a}_l\bm{a}_l^\transpose \\
	\bar{\mathbf{S}}_l &\gets (1-\lratePrecMod) \bar{\mathbf{S}}_l + \lratePrecMod \derivative{\bm{s}_l} {\derivative{\bm{s}_l}}^\transpose
\end{aligned}
\end{equation}
%We note that our scheme is not equivalent to performing natural gradient ascent directly on $\mathbf{A}_l$ and $\mathbf{S}_l$, so it is best regarded as a tractable approximation to eq.~\eqref{eq:ng-svi-G}.
Conveniently, because these factors are estimated from the empirical covariances, they (and hence also $\precision$) are automatically positive semidefinite.

Plugging the above formulas into eq.~\eqref{eq:ng-svi-G} does not quite yield an MVG posterior due to the addition of the prior Hessian.
In general, there may be no compact representation of $\precision$. However, for spherical Gaussian priors\footnote{We consider spherical Gaussian priors for simplicity, but this trick can be extended to any prior whose Hessian is Kronecker-factored, such as group sparsity.}, we can approximate $\Sigma$ using a trick proposed by \citet{martens2015optimizing} in the context of damping.
%In particular, we add $\pi_l \sqrt{\frac{\klWeight}{\ndata \priorVar}} \ident$ and $\frac{1}{\pi_l} \sqrt{\frac{\klWeight}{\ndata \priorVar}} \ident$ for a scalar constant $\pi_l$ to the individual Kronecker factors $\mathbf{A}_l$ and $\mathbf{S}_l$.
In this way, the covariance $\cov_l$ decomposes as the Kronecker product of two terms:
\begin{align}
\cov_l & = \frac{\klWeight}{\ndata} [\mathbf{S}_l^{\gamma}]^{-1} \otimes [\mathbf{A}_l^{\gamma}]^{-1} \label{eq:fisher-inv} \\
&\triangleq \frac{\klWeight}{\ndata} \left(\bar{\mathbf{S}}_l + \frac{1}{\pi_l} \sqrt{\frac{\klWeight}{\ndata \priorVar}} \ident \right)^{-1} \otimes \left(\bar{\mathbf{A}}_l + \pi_l \sqrt{\frac{\klWeight}{\ndata \priorVar}} \ident \right)^{-1} \nonumber
\end{align}
This factorization corresponds to a matrix-variate Gaussian posterior $\matrixNormal(\mathbf{W}_l ; \mathbf{M}_l, \frac{\klWeight}{N} [\mathbf{A}_l^{\gamma}]^{-1}, [\mathbf{S}_l^{\gamma}]^{-1})$, where the $\klWeight/N$ factor is arbitrarily assigned to the first factor. 
We refer to this BNN training method as \emph{noisy K-FAC}.
The full algorithm is given as Alg.~\ref{alg:noisy-kfac}. 

\begin{table*}[t]
\caption{Averaged test RMSE and log-likelihood for the regression benchmarks.}
\vspace{-1em}
\label{uci regression}
\begin{center}
\resizebox{\textwidth}{!}{
\begin{sc}
\begin{tabular}{lcccccccc}
\toprule
\textbf{}        & \multicolumn{4}{c}{Test RMSE} & \multicolumn{4}{c}{Test log-likelihood} \\
\textbf{Dataset} & BBB & PBP & NNG-FFG & NNG-MVG & BBB & PBP & NNG-FFG & NNG-MVG \\
\midrule
Boston  		 & 3.171$\pm$0.149 & 3.014$\pm$0.180 & 3.031$\pm$0.155 & \textbf{2.742$\pm$0.125} & -2.602$\pm$0.031 & -2.574$\pm$0.089 & -2.558$\pm$0.032 & \textbf{-2.446$\pm$0.029} \\
Concrete         & 5.678$\pm$0.087 & 5.667$\pm$0.093 & 5.613$\pm$0.113 & \textbf{5.019$\pm$0.127} & -3.149$\pm$0.018 & -3.161$\pm$0.019 & -3.145$\pm$0.023 & \textbf{-3.039$\pm$0.025} \\
Energy           & 0.565$\pm$0.018 & 1.804$\pm$0.048 & 0.839$\pm$0.046 & \textbf{0.485$\pm$0.023} & -1.500$\pm$0.006 & -2.042$\pm$0.019 & -1.629$\pm$0.020 & \textbf{-1.421$\pm$0.005} \\
Kin8nm           & 0.080$\pm$0.001 & 0.098$\pm$0.001 & 0.079$\pm$0.001 & \textbf{0.076$\pm$0.001} & 1.111$\pm$0.007 & 0.896$\pm$0.006 & 1.112$\pm$0.008 & \textbf{1.148$\pm$0.007} \\
Naval            & \textbf{0.000$\pm$0.000} & 0.006$\pm$0.000 & 0.001$\pm$0.000 & \textbf{0.000$\pm$0.000} & 6.143$\pm$0.032 & 3.731$\pm$0.006 & 6.231$\pm$0.041 & \textbf{7.079$\pm$0.034} \\
Pow. Plant       & 4.023$\pm$0.036 & 4.124$\pm$0.035 & 4.002$\pm$0.039 & \textbf{3.886$\pm$0.041} & -2.807$\pm$0.010 & -2.837$\pm$0.009 & -2.803$\pm$0.010 & \textbf{-2.776$\pm$0.011} \\
Protein          & 4.321$\pm$0.017 & 4.732$\pm$0.013 & 4.380$\pm$0.016 & \textbf{4.097$\pm$0.009} & -2.882$\pm$0.004 & -2.973$\pm$0.003 & -2.896$\pm$0.004 & \textbf{-2.836$\pm$0.002} \\
Wine             & 0.643$\pm$0.012 & \textbf{0.635$\pm$0.008} & 0.644$\pm$0.011 & 0.637$\pm$0.011 & -0.977$\pm$0.017 & \textbf{-0.968$\pm$0.014} & -0.976$\pm$0.016 & -0.969$\pm$0.014 \\
Yacht            & 1.174$\pm$0.086 & 1.015$\pm$0.054 & 1.289$\pm$0.069 & \textbf{0.979$\pm$0.077} & -2.408$\pm$0.007 & \textbf{-1.634$\pm$0.016} & -2.412$\pm$0.006 & -2.316$\pm$0.006 \\
Year             & 9.076$\pm$NA & \textbf{8.879$\pm$NA} & 9.071$\pm$NA & 8.885$\pm$NA & -3.614$\pm$NA & -3.603$\pm$NA & -3.620$\pm$NA & \textbf{-3.595$\pm$NA} \\ 
\bottomrule
\end{tabular}
\end{sc}
}
\end{center}
\end{table*}
\begin{algorithm}[t]
\caption{Noisy Adam. Differences from standard Adam are shown in {\color{blue} blue}.
}
\label{alg:noisy-adam}
\begin{algorithmic}
\REQUIRE $\alpha$: Stepsize
\REQUIRE $\beta_1, \beta_2$: Exponential decay rates for updating $\mean$ and the Fisher $\fisher$
\REQUIRE $\klWeight, \eta, \gamma_{\text{ex}}:$ KL weighting, prior variance, extrinsic damping term
\STATE $\mathbf{m} \leftarrow \mathbf{0}$
\STATE Calculate the intrinsic damping term ${\color{blue}\gamma_{\text{in}}=\frac{\klWeight}{\ndata \eta}}$, total damping term ${\color{blue}\gamma=\gamma_{\text{in}}+\gamma_{\text{ex}}}$
\WHILE{stopping criterion not met}
    \STATE {\color{blue} $\weights \sim \normal(\mathbf{\mean},\ \frac{\klWeight}{N}\diag (\mathbf{f} + \gamma_{\text{in}} )^{-1} )$} \phantom{$\mu^{(k)}$}
	\STATE $\mathbf{v} \leftarrow \nabla_{\weights}\log p(\target | \inputVec, \weights) {\color{blue} \,-\, \gamma_{\text{in}} \cdot \weights}$ \phantom{$\mu^{(k)}$}
	\STATE $\mathbf{m} \leftarrow \beta_1 \cdot \mathbf{m} + (1-\beta_1) \cdot \mathbf{v}$ \phantom{$\mu^{(k)}$} (Update momentum)
	\STATE $\mathbf{f} \leftarrow \beta_2 \cdot \mathbf{f} + (1 - \beta_2) \cdot (\nabla_{\weights}\log p(\target | \inputVec, \weights)^2$ \phantom{$\mu^{(k)}$}
	\STATE $\tilde{\mathbf{m}} \leftarrow \mathbf{m}/(1-\beta_1^k)$ \phantom{$\mu^{(k)}$}
	\STATE $\hat{\mathbf{m}} \leftarrow \tilde{\mathbf{m}} / {\color{blue}(\mathbf{f} + \gamma)}$  \phantom{$\mu^{(k)}$}
	\STATE $\mean \leftarrow \mean + \alpha \cdot \hat{\mathbf{m}}$ \hspace{1em} (Update parameters) \phantom{$\mu^{(k)}$}
\ENDWHILE
\end{algorithmic}
\end{algorithm}
\begin{algorithm}[t]
\caption{Noisy K-FAC.
Subscript $l$ denotes layers, $\weights_l = \kvec(\mathbf{W}_l)$, and $\mean_l = \kvec(\mathbf{M}_l)$.
We assume zero momentum for simplicity.
Differences from standard K-FAC are shown in {\color{blue} blue}.
}
\label{alg:noisy-kfac}
\begin{algorithmic}
\REQUIRE $\alpha$: stepsize
\REQUIRE $\beta$: exponential moving average parameter %for covariance factors
\REQUIRE $\klWeight, \eta, \gamma_{\text{ex}}:$ KL weighting, prior variance, extrinsic damping term
\REQUIRE stats and inverse update intervals $T_{\rm stats}$ and $T_{\rm inv}$
\STATE $k \leftarrow 0$ and initialize $\{\mean_l\}_{l=1}^{L}, \{\mathbf{S}_l\}_{l=1}^{L}, \{\mathbf{A}_l\}_{l=1}^{L}$
\STATE Calculate the intrinsic damping term ${\color{blue}\gamma_{\text{in}}=\frac{\klWeight}{\ndata \eta}}$, total damping term ${\color{blue}\gamma=\gamma_{\text{in}}+\gamma_{\text{ex}}}$
\WHILE{stopping criterion not met}
	\STATE $k \leftarrow k+1$
        \STATE {\color{blue} $\mathbf{W}_l \sim \matrixNormal(\mathbf{M}_l,\ \frac{\klWeight}{N} [\mathbf{A}_l^{\gamma_{\text{in}}}]^{-1},\ [\mathbf{S}_l^{\gamma_{\text{in}}}]^{-1})$}
        \IF {$k \equiv 0$ (\textrm{mod} $T_{\rm stats}$)}
%            \STATE Sample the targets from predictive distribution.
	    \STATE Update the factors $\{\mathbf{S}_l\}_{l=1}^{L}, \{\mathbf{A}_l\}_{l=0}^{L-1}$ using eq.~\eqref{eq:running-average}
        \ENDIF
        \IF {$k \equiv 0$ (\textrm{mod} $T_{\rm inv}$)}
	    \STATE Calculate the inverses $\{[\mathbf{S}_l^{\gamma}]^{-1}\}_{l=1}^{L}, \{[\mathbf{A}_l^{\gamma}]^{-1}\}_{l=0}^{L-1}$ using eq.~\eqref{eq:fisher-inv}.
        \ENDIF
        \STATE $\mathbf{V}_l = \nabla_{\mathbf{W}_l}\log p(\target | \inputVec, \weights) {\color{blue} \,-\, \gamma_{\text{in}} \cdot \mathbf{W}_l}$
        %\STATE $\mathbf{M}_l \gets \mathbf{M}_l + \alpha ([\mathbf{S}_l^{\gamma}]^{-1} \otimes [\mathbf{A}_l^{\gamma}]^{-1}) \mathbf{v}$
        \STATE $\mathbf{M}_l \gets \mathbf{M}_l + \alpha [\mathbf{A}_l^{\gamma}]^{-1} \mathbf{V}_l [\mathbf{S}_l^{\gamma}]^{-1}$
\ENDWHILE
\end{algorithmic}
\end{algorithm}

%K-FAC is a very efficient optimizer, and has been observed to yield significant speedups over standard methods for training convolutional networks~\citep{grosse2016kronecker}.
%%and improved sample efficiency in reinforcement learning~\citep{wu2017scalable}.
%The algorithm is GPU-friendly, and efficient implementations typically only introduce small (e.g. 1.5--2x) overhead compared with ordinary SGD.
%However, our focus here is not optimization speed, but rather the ability of noisy K-FAC to fit flexible MVG posteriors, in order to obtain improved uncertainty estimates compared with fully factorized Gaussians.

\subsection{Block Tridiagonal Covariance}
Both the fully factorized and MVG posteriors assumed independence between layers. However, in practice the weights in different layers can be tightly coupled.
To better capture these dependencies, we propose to approximate $\fisher$ using the block tridiagonal approximation from \citet{martens2015optimizing}.
The resulting posterior covariance is block tridiagonal, so it accounts for dependencies between adjacent layers.
The noisy version of block tridiagonal K-FAC is completely analogous to the block diagonal version, but since the approximation is rather complicated, we refer the reader to \citet{martens2015optimizing} for details.

\section{Related Work}
\label{sec:related}
Variational inference was first applied to neural networks by \citet{peterson1987mean} and \citet{hinton1993keeping}.
More recently, \citet{graves2011practical} proposed a practical method for variational inference with fully factorized Gaussian posteriors which used a simple (but biased) gradient estimator.
Improving on that work, \citet{blundell2015weight} proposed a unbiased gradient estimator using the reparameterization trick of \citet{kingma2013auto}.
%In place of the ELBO, \citet{hernandez2015probabilistic} proposed to use expectation propagation with fully factorized posterior distributions and found a closed-form approximation which avoided the sampling for weights.
\citet{kingma2015variational} observed that variance of stochastic gradients can be significantly reduced by local reparameterization trick where global uncertainty in the weights is translated into local uncertainty in the activations.

There has also been much work on modeling the correlations between weights using more complex Gaussian variational posteriors.
\citet{louizos2016structured} introduced the matrix variate Gaussian posterior as well as a Gaussian process approximation.
\citet{sun2017learning} decoupled the correlations of a matrix variate Gaussian posterior to unitary transformations and factorial Gaussian.
Inspired by the idea of normalizing flows in latent variable models~\citep{rezende2015variational}, \citet{louizos2017multiplicative} applied normalizing flows to auxiliary latent variables to produce more flexible approximate posteriors.
%However, the introduction of auxiliary variables can result a looser variational lower bound.

Since natural gradient was proposed by \citet{amari1998natural}, there has been much work on tractable approximations.
\citet{hoffman2013stochastic} observed that for exponential family posteriors, the exact natural gradient could be tractably computed using stochastic versions of variational Bayes E-M updates.
\citet{martens2015optimizing} proposed K-FAC for performing efficient natural gradient optimization in deep neural networks.
Following on that work, K-FAC has been adopted in many tasks~\citep{grosse2016kronecker, wu2017scalable} to gain optimization benefits, and was shown to be amenable to distributed computation~\citep{ba2017distributed}.

\citet{khan2017variational} independently derived a stochastic Newton update similar to eq.~\eqref{eq:ng-update}.
Their focus was on variational optimization (VO)~\citep{staines2012variational} which can be seen as a special case of variational inference by omitting KL term, and they only derived the version of diagonal approximation (see Section~\ref{sec:ffg}).  
Assuming the variational distribution is Gaussian distribution, we can extend NNG to VO by only modifying the update rule of Fisher to keep a running sum of individual Fisher.

Concurrently, \citet{khan2017vprop} also found the relationship between natural gradient and variational inference and derived Vprop by adding weight noise to RMSprop, which essentially resembles noisy Adam.
\section{Experiments}
In this section, we conducted a series of experiments to investigate the following questions: (1) How does noisy natural gradient (NNG) compare with existing methods in terms of prediction performance? (2) Is NNG able to scale to large dataset and modern-size convolutional neural network? (3) Can NNG achieve better uncertainty estimates? (4) Does it enable more efficient exploration in active learning and reinforcement learning?

%To evaluate the effectiveness of our method, we compared it with Bayes By Backprop (BBB)~\citep{blundell2015weight}, probabilistic backpropagation (PBP) with a factorial gaussian posterior~\citep{hernandez2015probabilistic}.
Our method with a full-covariance multivariate Gaussian, a fully-factorized Gaussian, a matrix-variate Gaussian and block-tridiagonal posterior are denoted as NNG-full, NNG-FFG~(noise Adam), NNG-MVG~(noisy K-FAC) and NNG-BlkTri, respectively.

\subsection{Regression}
We first experimented with regression datasets from the UCI collection~\citep{asuncion2007uci}.
All experiments used networks with one hidden layer unless stated otherwise. We compared our method with Bayes By Backprop (BBB)~\citep{blundell2015weight}, probabilistic backpropagation (PBP) with a factorial gaussian posterior~\citep{hernandez2015probabilistic}. The results for PBP\_MV~\citep{sun2017learning} and VMG~\citep{louizos2016structured} can be found in supplement. 

Following previous work~\citep{hernandez2015probabilistic,louizos2016structured}, we report the standard metrics including root mean square error (RMSE) and test log-likelihood.
The results are summarized in Table~\ref{uci regression}.
As we can see from the results, our NNG-FFG performed similarly to BBB~\citep{blundell2015weight}, indicating that the Graves approximation did not cause a performance hit.
Our NNG-MVG method achieved substantially better RMSE and log-likelihoods than BBB and PBP due to the more flexible posterior. Moreover, NNG-MVG outperforms PBP\_MV~\citep{sun2017learning} on all datasets other than Yacht and Year, though PBP\_MV also uses matrix variate Gaussian posterior.

%\begin{figure}[ht]
%  \centering
%  {\includegraphics[width=\linewidth]{imgs/comparision.pdf}}
%  \caption{Training curves for all three methods. For each method, we tuned the learning rate for updating the posterior mean. Note that BBB and NNG-FFG use the same form of $q$, while NNG-MVG uses a more flexible $q$ distribution.
%\label{fig:convergence}}
%\end{figure}
%
%
%While optimization was not the primary focus of this work, we compared NNG with the baseline BBB in terms of convergence
%(Training curves for two regression datasets can be found in supplement).
%We found that NNG-FFG trained in fewer iterations than BBB, while leveling off to similar ELBO values, even though our BBB implementation used Adam, and hence itself exploited diagonal curvature.
%Furthermore, despite the increased flexibility and larger number of parameters, NNG-MVG took roughly 2 times fewer iterations to converge, while at the same time surpassing BBB by a significant margin in terms of the ELBO.
\begin{table}[t]
\caption{Classification accuracy on CIFAR10 with modified VGG16. {\bf [D]} denotes data augmentation including horizontal flip and random crop while {\bf [B]} denotes Batch Normalization. We leave {\bf [N/A]} for BBB and noisy Adam with BN since they are extremely unstable and work only with a very small $\klWeight$.}
\vspace{-1em}
\label{tab:classification}
\begin{center}
\resizebox{0.9\columnwidth}{!}{
\begin{sc}
\begin{tabular}{l|c|c|c|c|c}
\toprule
\multirow{2}{*}{Method} & \multirow{2}{*}{Network} & \multicolumn{4}{c}{Test Accuracy}  \\ 
\cline{3-6}
&  &  &  D & B &  D + B \\
\midrule
SGD         &  VGG16  & 81.79 & 88.35 & 85.75 & 91.39 \\
KFAC        &  VGG16  & 82.39 & 88.89 & 86.86 & \textbf{92.13} \\
BBB         &  VGG16  & 82.82 & 88.31 & N/A  & N/A \\
Noisy-Adam  &  VGG16  & 82.68 & 88.23 & N/A  & N/A \\
Noisy-KFAC  &  VGG16  & \textbf{85.52} & \textbf{89.35} & \textbf{88.22} & 92.01 \\
\bottomrule
\end{tabular}
\end{sc}
}
\end{center}
\vspace{-1em}
\end{table}
\begin{table*}[t]
\caption{Average test RMSE in active learning.}
\vspace{-1em}
\label{tab:active learning}
\begin{center}
\resizebox{\textwidth}{!}{
\begin{sc}
\begin{tabular}{lcccccccc}
\toprule
\textbf{Dataset} & PBP\_R & PBP\_A & NNG-FFG\_R & NNG-FFG\_A & NNG-MVG\_R & NNG-MVG\_A & HMC\_R & HMC\_A\\
\midrule
Boston  		 & 6.716$\pm$0.500 & 5.480$\pm$0.175 & 5.911$\pm$0.250 & 5.435$\pm$0.132 & 5.831$\pm$0.177 & 5.220$\pm$0.132 & 5.750$\pm$0.222 & \textbf{5.156$\pm$0.150} \\
Concrete         & 12.417$\pm$0.392 & 11.894$\pm$0.254 & 12.583$\pm$0.168 & 12.563$\pm$0.142 & 12.301$\pm$0.203 & 11.671$\pm$0.175 & \textbf{10.564$\pm$0.198} & 11.484$\pm$0.191 \\
Energy           & 3.743$\pm$0.121 & 3.399$\pm$0.064 & 4.011$\pm$0.087 & 3.761$\pm$0.068 & 3.635$\pm$0.084 & 3.211$\pm$0.076 & 3.264$\pm$0.067 & \textbf{3.118$\pm$0.062} \\
Kin8nm           & 0.259$\pm$0.006 & 0.254$\pm$0.005 & 0.246$\pm$0.004 & 0.252$\pm$0.003 & 0.243$\pm$0.003 & 0.244$\pm$0.003 & 0.226$\pm$0.004 & \textbf{0.223$\pm$0.003} \\
Naval            & 0.015$\pm$0.000 & 0.016$\pm$0.000 & 0.013$\pm$0.000 & 0.013$\pm$0.000 & 0.010$\pm$0.000 & \textbf{0.009$\pm$0.000} & 0.013$\pm$0.000 & 0.012$\pm$0.000 \\
Pow. Plant       & 5.312$\pm$0.108 & 5.068$\pm$0.082 & 5.812$\pm$0.119 & 5.423$\pm$0.111 & 5.377$\pm$0.133 & 4.974$\pm$0.078 & 5.229$\pm$0.097 & \textbf{4.800$\pm$0.074} \\
Wine             & 0.945$\pm$0.044 & 0.809$\pm$0.011 & 0.730$\pm$0.011 & 0.748$\pm$0.008 & 0.752$\pm$0.014 & 0.746$\pm$0.009 & \textbf{0.740$\pm$0.011} & 0.749$\pm$0.010 \\
Yacht            & 5.388$\pm$0.339 & 4.508$\pm$0.158 & 7.381$\pm$0.309 & 6.583$\pm$0.264 & 7.192$\pm$0.280 & 6.371$\pm$0.204 & 4.644$\pm$0.237 & \textbf{3.211$\pm$0.120} \\
\bottomrule
\end{tabular}
\end{sc}
}
\end{center}
\end{table*}
\subsection{Classification}
To evaluate the scalability of our method to large networks, we applied noisy K-FAC to a modified version of the VGG16\footnote{The detailed network architecture is 32-32-M-64-64-M-128-128-128-M-256-256-256-M-256-256-256-M-FC10, where each number represents the number of filters in a convolutional layer, and M denotes max-pooling.} network~\citep{simonyan2014very} and tested it on CIFAR10 benchmark~\citep{krizhevsky2009learning}. 
It is straightforward to incorporate noisy K-FAC into convolutional layers by considering them using Kronecker Factors for Convolution \citet{grosse2016kronecker}. 
We compared our method to SGD with momentum, K-FAC and BBB in terms of test accuracy.
%We found the adaptive weight noise can be interpreted as regularization term which helps resist overfitting and thus improving classification accuracy.
Results are shown in Table~\ref{tab:classification}. Noisy K-FAC achieves the highest accuracy on all configurations except where both data augmentation and Batch Normalization (BN)~\citep{ioffe2015batch} are used. When no extra regularization used, noisy K-FAC shows a gain of 3\% (85.52\% versus 82.39\%). 

\begin{figure}[h]
  \centering
  {\includegraphics[width=0.9\linewidth]{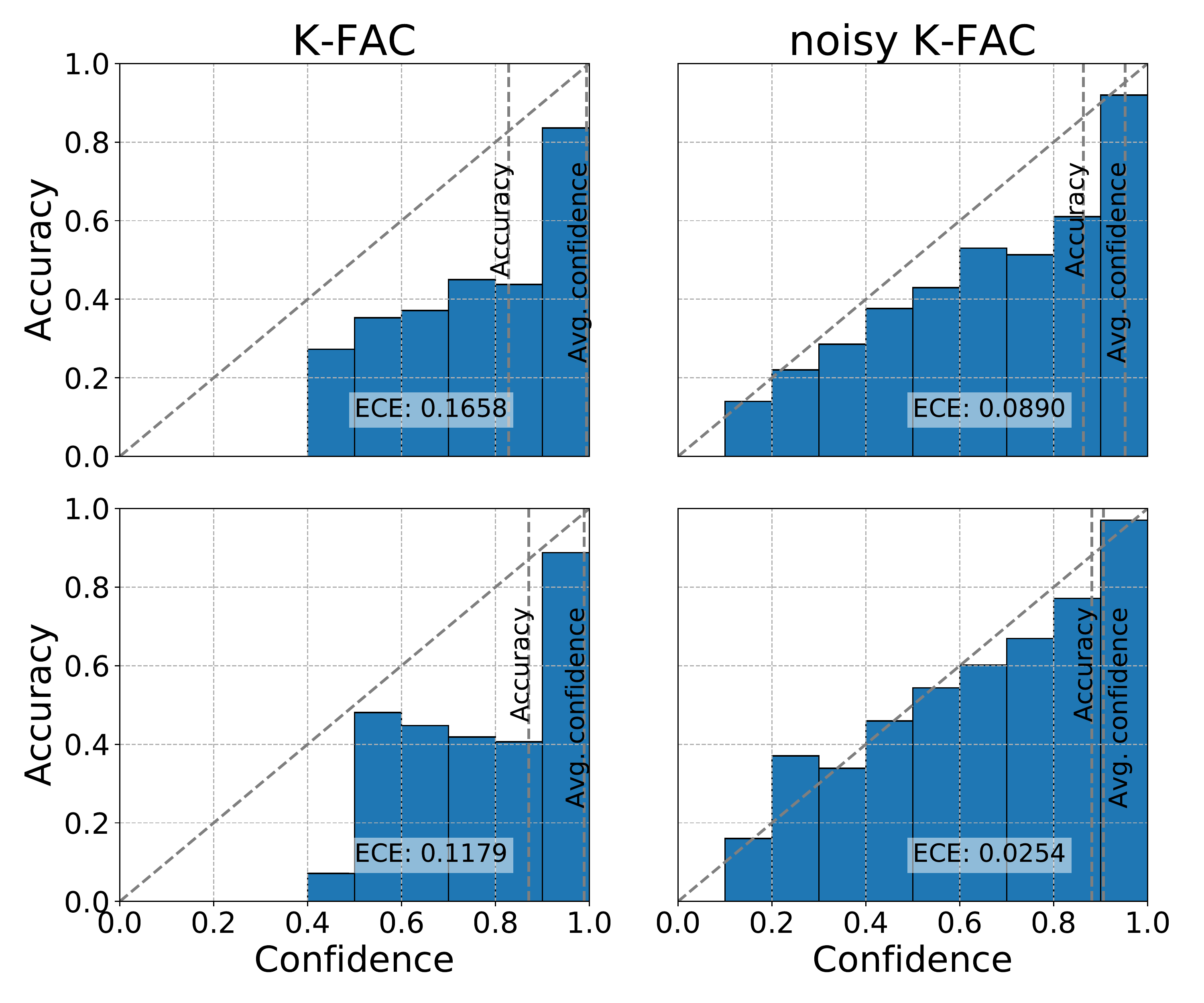}}
  \vspace{-1em}
  \caption{Reliability diagrams \cite{niculescu2005predicting,guo2017calibration} for K-FAC (left) and noisy K-FAC (right) on CIFAR10. Reliability diagrams show accuracy as a function of confidence. Models trained without BN (top) and with BN (bottom). ECE = Expected Calibration Error \cite{guo2017calibration}; smaller is better.
\label{fig:model_calibration}}
\end{figure}

We observed that point estimates tend to make poorly calibrated predictions, as shown in Figure~\ref{fig:model_calibration}.
By contrast, models trained with noisy K-FAC are well-calibrated (i.e. the bars align roughly along the diagonal), which benefits interpretability. 
%We observed that point estimates trained with BN tend to make poorly calibrated predictions. 
%By contrast, models trained with noisy K-FAC are well-calibrated, which benefits interpretability. 
%As shown in Figure~\ref{fig:model_calibration} that BN can be naturally incorporated into noisy K-FAC in which case well-calibrated confidence estimates are retained (i.e. the bars align roughly along the diagonal).

We note that noisy K-FAC imposes a weight decay term intrinsically. To check that this by itself doesn't explain the performance gains, we modified K-FAC to use weight decay of the same magnitude. K-FAC with this weight decay setting achieves 83.51\% accuracy. However, as shown in Table~\ref{tab:classification}, noisy K-FAC achieves 85.52\%, demonstrating the importance of adaptive weight noise.

\subsection{Active Learning}
One particularly promising application of uncertainty estimation is to guiding an agent's exploration towards part of a space which it's most unfamiliar with. We have evaluated our BNN algorithms in two instances of this general approach: active learning, and intrinsic motivation for reinforcement learning. The next two sections present experiments in these two domains, respectively.

In the simplest active learning setting~\citep{settles2010active}, an algorithm is given a set of unlabeled examples and, in each round, chooses one unlabeled example to have labeled.
A classic Bayesian approach to active learning is the information gain criterion~\citep{mackay1992information}, which in each step attempts to achieve the maximum reduction in posterior entropy.
Under the assumption of i.i.d.\ Gaussian noise, this is equivalent to choosing the unlabeled example with the largest predictive variance.
%Active learning using the information gain criterion was introduced as a BNN benchmark by \citet{hernandez2015probabilistic}; our experiments are based on their protocol.
%All methods under consideration use the same neural network architecture and prior, and differ only in the inference procedure.
\begin{table}[h]
\caption{Pearson correlation of each algorithm's predictive variances with those of HMC.}
\vspace{-1em}
\label{tab:act-corr}
\begin{center}
\resizebox{\columnwidth}{!}{
\begin{sc}
\begin{tabular}{lcccc}
\toprule
Dataset    & PBP              & NNG-FFG            & NNG-MVG            & NNG-BlkTri      \\
\midrule
Boston     & 0.761$\pm$0.032  & 0.718$\pm$0.035    & \textbf{0.891$\pm$0.021}  & 0.889$\pm$0.024   \\
Concrete   & 0.817$\pm$0.028  & 0.811$\pm$0.028    & 0.913$\pm$0.010    & \textbf{0.922$\pm$0.006} \\
Energy     & 0.471$\pm$0.076  & 0.438$\pm$0.075    & 0.617$\pm$0.087    & \textbf{0.646$\pm$0.088} \\
Kin8nm     & 0.587$\pm$0.021  & 0.659$\pm$0.015    & 0.731$\pm$0.021    & \textbf{0.759$\pm$0.023} \\
Naval      & 0.270$\pm$0.098  & 0.321$\pm$0.087    & 0.596$\pm$0.073    & \textbf{0.598$\pm$0.070} \\
Pow. Plant & 0.509$\pm$0.068  & 0.618$\pm$0.050    & 0.829$\pm$0.020    & \textbf{0.853$\pm$0.020} \\
Wine       & 0.883$\pm$0.042  & 0.918$\pm$0.014    & 0.957$\pm$0.009    & \textbf{0.964$\pm$0.006} \\
Yacht      & 0.620$\pm$0.053  & 0.597$\pm$0.063    & 0.717$\pm$0.072    & \textbf{0.727$\pm$0.070} \\
\bottomrule
\end{tabular}
\end{sc}
}
\end{center}
\end{table}
\begin{figure*}[t]
\centering
\subfigure[CartPoleSwingup] { \label{fig:rl-trpo-cartpole} 
\includegraphics[height=3cm]{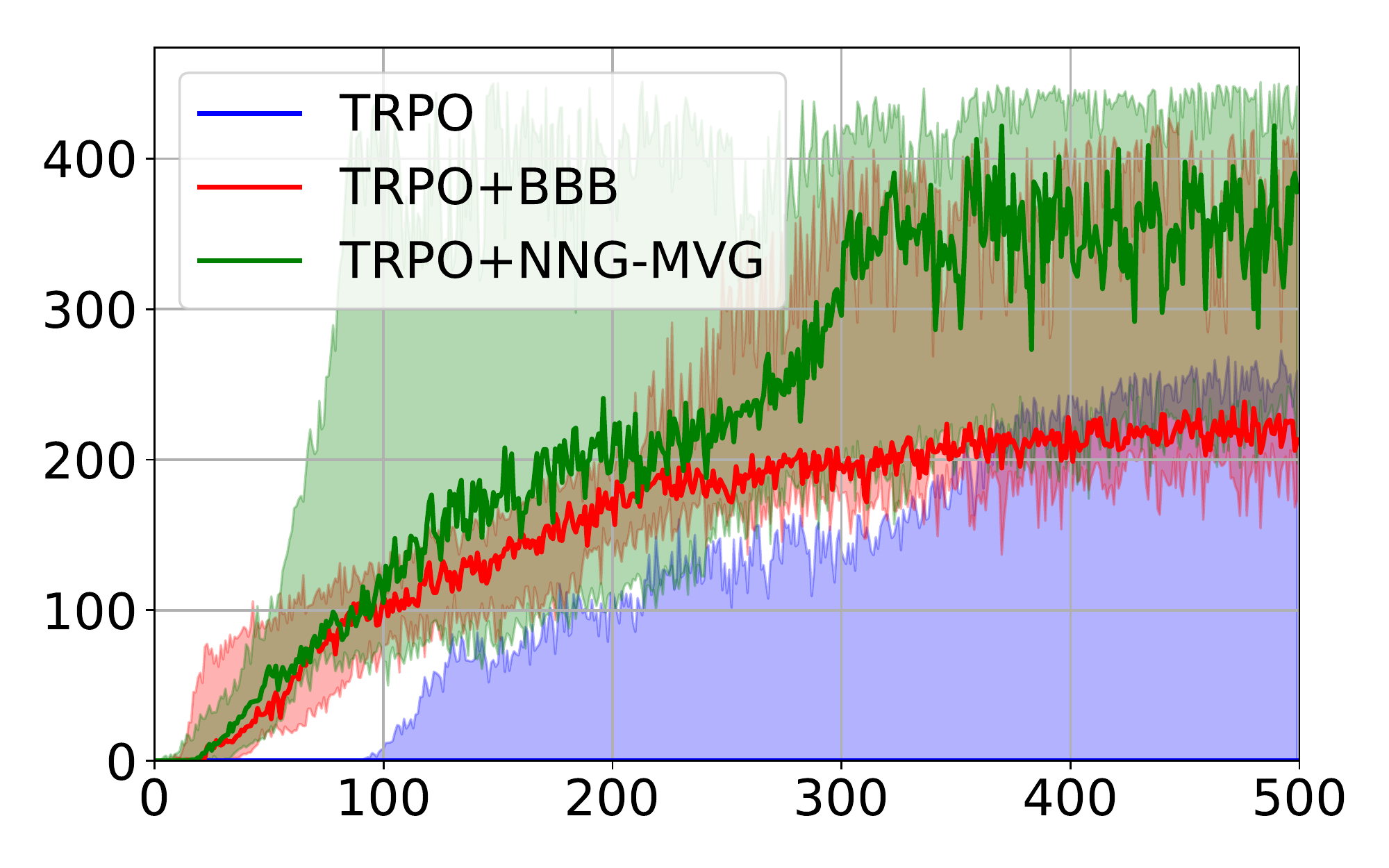} 
} 
\subfigure[MountainCar] { \label{fig:rl-trpo-mountain} 
\includegraphics[height=3cm]{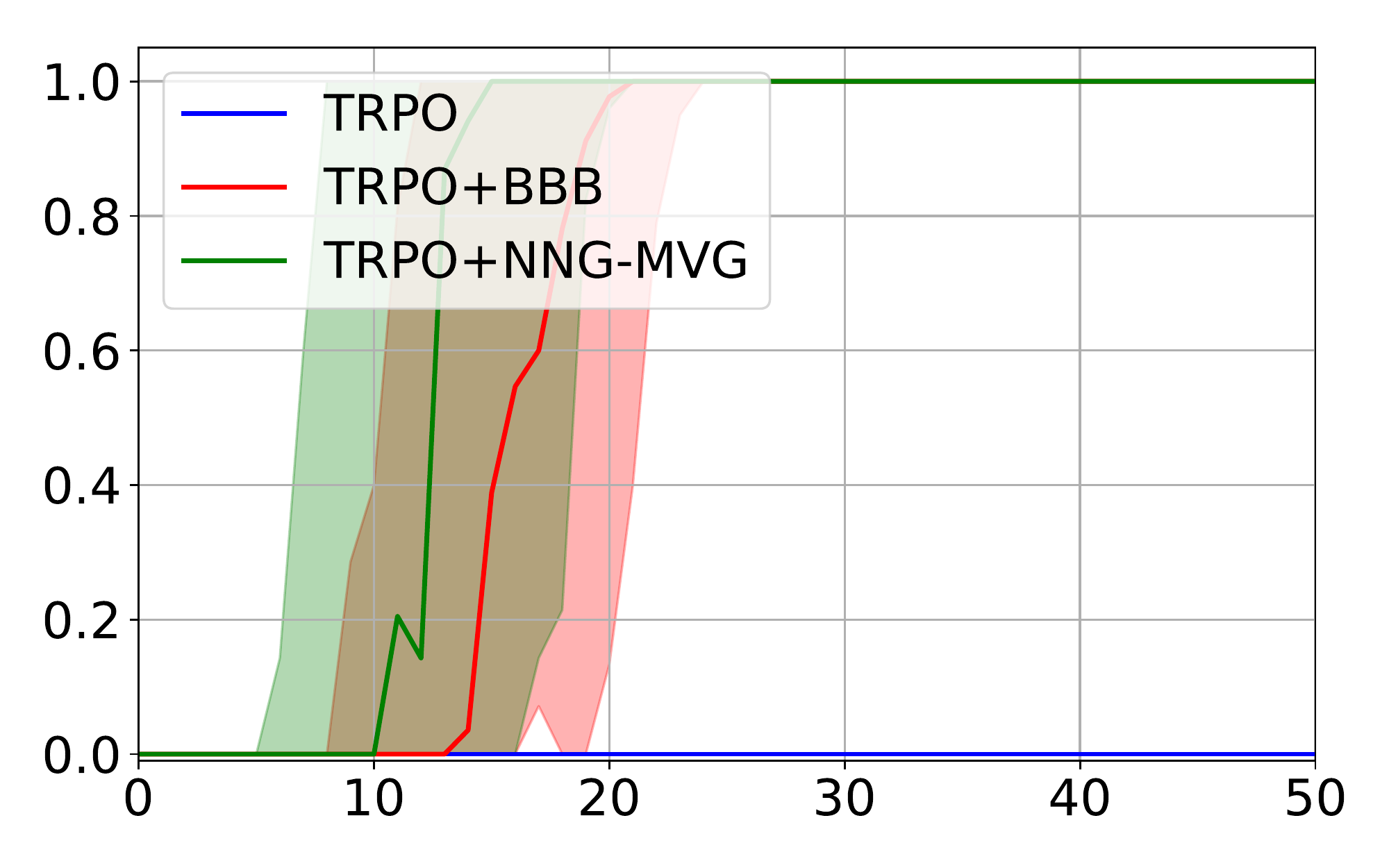} 
}
\subfigure[DoublePendulum] { \label{fig:rl-trpo-pendulum} 
\includegraphics[height=3cm]{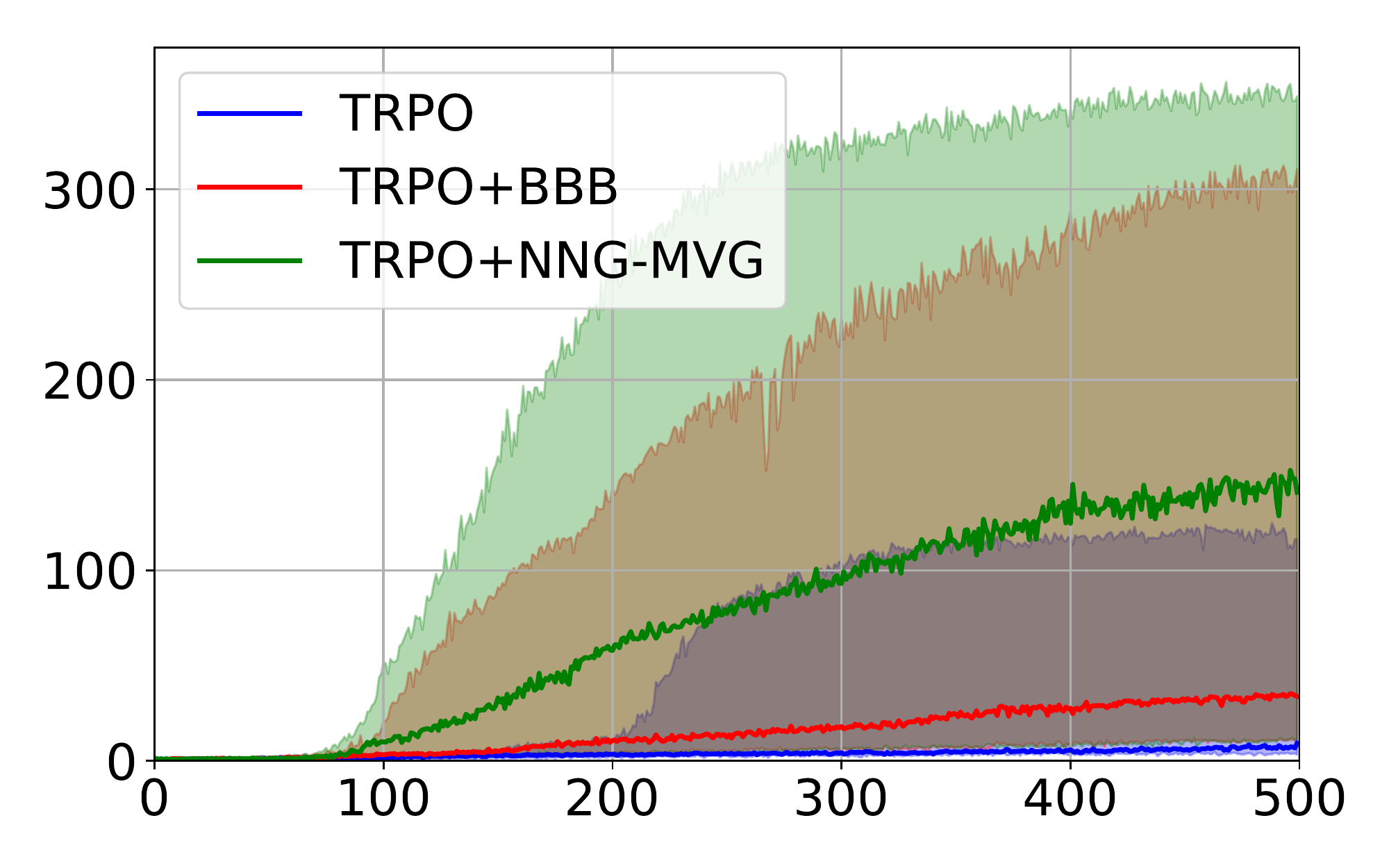} 
} 
\vspace{-1em}
\caption{Performance of {\bf [TRPO]} TRPO baseline with Gaussian control noise, {\bf [TRPO+BBB]} VIME baseline with BBB dynamics network, and {\bf [TRPO+NNG-MVG]} VIME with NNG-MVG dynamics network (ours).
The darker-colored lines represent the median performance in 10 different random seeds while the shaded area show the interquartile range.} 
\label{fig:rl} 
\end{figure*}
We first investigated how accurately each of the algorithms could estimate predictive variances.
In each trial, we randomly selected 20 labeled training examples and 100 unlabeled examples; we then computed each algorithm's posterior predictive variances for the unlabeled examples.
10 independent trials were run.
As is common practice, we treated the predictive variance of HMC as the ``ground truth'' predictive variance.
Table~\ref{tab:act-corr} reports the average and standard error of Pearson correlations between the predictive variances of each algorithm and those of HMC.
In all of the datasets, our two methods NNG-MVG and NNG-BlkTri match the HMC predictive variances significantly better than the other approaches, and NNG-BlkTri consistently matches them slightly better than NNG-MVG due to the more flexible variational posterior.

Next, we evaluated the performance of all methods on active learning, following the protocol of \citet{hernandez2015probabilistic}.
As a control, we Ωevaluated each algorithm with labeled examples selected uniformly at random; this is denoted with the \_R suffix.
Active learning results are denoted with the \_A suffix.
The average test RMSE for all methods is reported in Table \ref{tab:active learning}.
%The average test RMSE for all methods can be found in supplement.
These results shows that NNG-MVG\_A performs better than NNG-MVG\_R in most datasets and is closer to HMC\_A compared to PBP\_A and NNG-FFG\_A.
However, we note that better predictive variance estimates do not reliably yield better active learning results, and in fact, active learning methods sometimes perform worse than random.
Therefore, while information gain is a useful criterion for benchmarking purposes, it is important to explore other uncertainty-based active learning criteria.
\subsection{Reinforcement Learning}
We next experimented with using uncertainty to provide intrinsic motivation in reinforcement learning.
%Reinforcement learning problems with sparse rewards can be particularly challenging, since the agent may need to execute a complex behavior even to obtain a single nonzero reward. 
\citet{houthooft2016vime} proposed Variational Information Maximizing Exploration (VIME), which encouraged the agent to seek novelty through an information gain criterion. 
VIME involves training a separate BNN to predict the dynamics, i.e.~learn to model the distribution $p(s_{t+1}|s_t, a_t; \theta)$.
With the idea that surprising states lead to larger updates to the dynamics network, the reward function was augmented with an ``intrinsic term'' corresponding to the information gain for the BNN. 
If the history of the agent up until time step $t$ is denoted as $\xi = \{s_1, a_1, ..., s_t\}$, then the modified reward can be written in the following form:
\begin{align}
r^*(s_t, a_t, s_{t+1}) & = r(s_t, a_t) \nonumber \\
& + \eta \kldiv (p(\theta|\xi_t, a_t, s_{t+1}) \klbars p(\theta| \xi_t))
\label{eq:reward}
\end{align}
%
%where $r(s_t,a_t)$ is the original reward function or external reward, and $\eta \in \mathbb{R}_{+}$ is a hyperparameter controlling the urge to explore.
In the above formulation, the true posterior is generally intractable.
\citet{houthooft2016vime} approximated it using Bayes by Backprop (BBB)~\citep{blundell2015weight}.
We experimented with replacing the fully factorized posterior with our NNG-MVG model.

Following the experimental setup of \citet{houthooft2016vime}, we tested our method in three continuous control tasks and sparsified the rewards (see supplement for details).
We compared our NNG-MVG dynamics model with a Gaussian noise baseline, as well as the original VIME formulation using BBB.
All experiments used TRPO to optimize the policy itself~\citep{schulman2015trust}.
%Because all the methods used the same policy gradient method and only differed only in the dynamics model, the performance differences should reflect the effectiveness of the BNN's uncertainty measure for detecting novelty.

Performance is measured by the average return (under the original MDP's rewards, not including the intrinsic term) at each iteration.
Figure~\ref{fig:rl} shows the performance results in three tasks. Consistently with \citet{houthooft2016vime}, we observed that the Gaussian noise baseline completely breaks down and rarely achieves the goal, VIME significantly improved the performance.
However, replacing the dynamics network with NNG-MVG considerably improved the exploration efficiency on all three tasks.
Since the policy search algorithm was shared between all three conditions, we attribute this improvement to the improved uncertainty modeling by the dynamics network.

\section{Conclusion}
We drew a surprising connection between natural gradient ascent for point estimation and for variational inference.
We exploited this connection to derive surprisingly simple variational BNN training procedures which can be instantiated as noisy versions of widely used optimization algorithms for point estimation.
This let us efficiently fit MVG variational posteriors, which capture correlations between different weights.
Our variational BNNs with MVG posteriors matched the predictive variances of HMC much better than fully factorized posteriors, and led to more efficient exploration in the settings of active learning and reinforcement learning with intrinsic motivation.

\section*{Acknowledgements}
GZ was supported by an NSERC Discovery Grant, and SS was supported by a Connaught New Researcher Award and a Connaught Fellowship. 
We thank Emtiyaz Khan and Mark van der Wilk for helpful discussions.

\bibliography{icml2018}

\begin{thebibliography}{44}
\providecommand{\natexlab}[1]{#1}
\providecommand{\url}[1]{\texttt{#1}}
\expandafter\ifx\csname urlstyle\endcsname\relax
  \providecommand{\doi}[1]{doi: #1}\else
  \providecommand{\doi}{doi: \begingroup \urlstyle{rm}\Url}\fi

\bibitem[Amari(1997)]{amari1997neural}
Amari, Shun-ichi.
\newblock Neural learning in structured parameter spaces-natural riemannian
  gradient.
\newblock In \emph{Advances in neural information processing systems}, pp.\
  127--133, 1997.

\bibitem[Amari(1998)]{amari1998natural}
Amari, Shun-Ichi.
\newblock Natural gradient works efficiently in learning.
\newblock \emph{Neural computation}, 10\penalty0 (2):\penalty0 251--276, 1998.

\bibitem[Asuncion \& Newman(2007)Asuncion and Newman]{asuncion2007uci}
Asuncion, Arthur and Newman, David.
\newblock Uci machine learning repository, 2007.

\bibitem[Ba et~al.(2016)Ba, Grosse, and Martens]{ba2016distributed}
Ba, Jimmy, Grosse, Roger, and Martens, James.
\newblock Distributed second-order optimization using kronecker-factored
  approximations.
\newblock 2016.

\bibitem[Ba et~al.(2017)Ba, Martens, and Grosse]{ba2017distributed}
Ba, Jimmy, Martens, James, and Grosse, Roger.
\newblock Distributed second-order optimization using kronecker-factored
  approximations.
\newblock In \emph{International Conference on Learning Representations}, 2017.

\bibitem[Blundell et~al.(2015)Blundell, Cornebise, Kavukcuoglu, and
  Wierstra]{blundell2015weight}
Blundell, Charles, Cornebise, Julien, Kavukcuoglu, Koray, and Wierstra, Daan.
\newblock Weight uncertainty in neural networks.
\newblock \emph{arXiv preprint arXiv:1505.05424}, 2015.

\bibitem[Bonnet(1964)]{bonnet1964transformations}
Bonnet, Georges.
\newblock Transformations des signaux al{\'e}atoires a travers les systemes non
  lin{\'e}aires sans m{\'e}moire.
\newblock \emph{Annals of Telecommunications}, 19\penalty0 (9):\penalty0
  203--220, 1964.

\bibitem[Duchi et~al.(2011)Duchi, Hazan, and Singer]{duchi2011adaptive}
Duchi, John, Hazan, Elad, and Singer, Yoram.
\newblock Adaptive subgradient methods for online learning and stochastic
  optimization.
\newblock \emph{Journal of Machine Learning Research}, 12:\penalty0 2121--2159,
  2011.

\bibitem[Graves(2011)]{graves2011practical}
Graves, Alex.
\newblock Practical variational inference for neural networks.
\newblock In \emph{Advances in Neural Information Processing Systems}, pp.\
  2348--2356, 2011.

\bibitem[Grosse \& Martens(2016)Grosse and Martens]{grosse2016kronecker}
Grosse, Roger and Martens, James.
\newblock A kronecker-factored approximate fisher matrix for convolution
  layers.
\newblock In \emph{International Conference on Machine Learning}, pp.\
  573--582, 2016.

\bibitem[Guo et~al.(2017)Guo, Pleiss, Sun, and Weinberger]{guo2017calibration}
Guo, Chuan, Pleiss, Geoff, Sun, Yu, and Weinberger, Kilian~Q.
\newblock On calibration of modern neural networks.
\newblock \emph{arXiv preprint arXiv:1706.04599}, 2017.

\bibitem[Hern{\'a}ndez-Lobato \& Adams(2015)Hern{\'a}ndez-Lobato and
  Adams]{hernandez2015probabilistic}
Hern{\'a}ndez-Lobato, Jos{\'e}~Miguel and Adams, Ryan.
\newblock Probabilistic backpropagation for scalable learning of bayesian
  neural networks.
\newblock In \emph{International Conference on Machine Learning}, pp.\
  1861--1869, 2015.

\bibitem[Hinton \& Van~Camp(1993)Hinton and Van~Camp]{hinton1993keeping}
Hinton, Geoffrey~E and Van~Camp, Drew.
\newblock Keeping the neural networks simple by minimizing the description
  length of the weights.
\newblock In \emph{Proceedings of the sixth annual conference on Computational
  learning theory}, pp.\  5--13. ACM, 1993.

\bibitem[Hoffman et~al.(2013)Hoffman, Blei, Wang, and
  Paisley]{hoffman2013stochastic}
Hoffman, Matthew~D, Blei, David~M, Wang, Chong, and Paisley, John.
\newblock Stochastic variational inference.
\newblock \emph{The Journal of Machine Learning Research}, 14\penalty0
  (1):\penalty0 1303--1347, 2013.

\bibitem[Houthooft et~al.(2016)Houthooft, Chen, Duan, Schulman, De~Turck, and
  Abbeel]{houthooft2016vime}
Houthooft, Rein, Chen, Xi, Duan, Yan, Schulman, John, De~Turck, Filip, and
  Abbeel, Pieter.
\newblock Vime: Variational information maximizing exploration.
\newblock In \emph{Advances in Neural Information Processing Systems}, pp.\
  1109--1117, 2016.

\bibitem[Ioffe \& Szegedy(2015)Ioffe and Szegedy]{ioffe2015batch}
Ioffe, Sergey and Szegedy, Christian.
\newblock Batch normalization: Accelerating deep network training by reducing
  internal covariate shift.
\newblock In \emph{International conference on machine learning}, pp.\
  448--456, 2015.

\bibitem[Khan et~al.(2017{\natexlab{a}})Khan, Lin, Tangkaratt, Liu, and
  Nielsen]{khan2017variational}
Khan, Mohammad~Emtiyaz, Lin, Wu, Tangkaratt, Voot, Liu, Zouzhu, and Nielsen,
  Didrik.
\newblock Variational adaptive-{Newton} method for explorative learning.
\newblock \emph{arXiv preprint arXiv:1711.05560}, 2017{\natexlab{a}}.

\bibitem[Khan et~al.(2017{\natexlab{b}})Khan, Liu, Tangkaratt, and
  Gal]{khan2017vprop}
Khan, Mohammad~Emtiyaz, Liu, Zuozhu, Tangkaratt, Voot, and Gal, Yarin.
\newblock Vprop: Variational inference using rmsprop.
\newblock \emph{arXiv preprint arXiv:1712.01038}, 2017{\natexlab{b}}.

\bibitem[Kingma \& Ba(2014)Kingma and Ba]{kingma2014adam}
Kingma, Diederik and Ba, Jimmy.
\newblock Adam: A method for stochastic optimization.
\newblock \emph{arXiv preprint arXiv:1412.6980}, 2014.

\bibitem[Kingma \& Welling(2013)Kingma and Welling]{kingma2013auto}
Kingma, Diederik~P and Welling, Max.
\newblock Auto-encoding variational bayes.
\newblock \emph{arXiv preprint arXiv:1312.6114}, 2013.

\bibitem[Kingma et~al.(2015)Kingma, Salimans, and
  Welling]{kingma2015variational}
Kingma, Diederik~P, Salimans, Tim, and Welling, Max.
\newblock Variational dropout and the local reparameterization trick.
\newblock In \emph{Advances in Neural Information Processing Systems}, pp.\
  2575--2583, 2015.

\bibitem[Krizhevsky(2009)]{krizhevsky2009learning}
Krizhevsky, Alex.
\newblock Learning multiple layers of features from tiny images.
\newblock 2009.

\bibitem[Louizos \& Welling(2016)Louizos and Welling]{louizos2016structured}
Louizos, Christos and Welling, Max.
\newblock Structured and efficient variational deep learning with matrix
  gaussian posteriors.
\newblock In \emph{International Conference on Machine Learning}, pp.\
  1708--1716, 2016.

\bibitem[Louizos \& Welling(2017)Louizos and
  Welling]{louizos2017multiplicative}
Louizos, Christos and Welling, Max.
\newblock Multiplicative normalizing flows for variational bayesian neural
  networks.
\newblock \emph{arXiv preprint arXiv:1703.01961}, 2017.

\bibitem[MacKay(1992{\natexlab{a}})]{mackay1992information}
MacKay, David~JC.
\newblock Information-based objective functions for active data selection.
\newblock \emph{Neural computation}, 4\penalty0 (4):\penalty0 590--604,
  1992{\natexlab{a}}.

\bibitem[MacKay(1992{\natexlab{b}})]{mackay1992practical}
MacKay, David~JC.
\newblock A practical {Bayesian} framework for backpropagation networks.
\newblock \emph{Neural Computation}, 4:\penalty0 448--472, 1992{\natexlab{b}}.

\bibitem[Martens(2014)]{martens2014new}
Martens, James.
\newblock New insights and perspectives on the natural gradient method.
\newblock \emph{arXiv preprint arXiv:1412.1193}, 2014.

\bibitem[Martens \& Grosse(2015)Martens and Grosse]{martens2015optimizing}
Martens, James and Grosse, Roger.
\newblock Optimizing neural networks with kronecker-factored approximate
  curvature.
\newblock In \emph{International Conference on Machine Learning}, pp.\
  2408--2417, 2015.

\bibitem[Neal(1995)]{neal1995bayesian}
Neal, Radford~M.
\newblock \emph{BAYESIAN LEARNING FOR NEURAL NETWORKS}.
\newblock PhD thesis, University of Toronto, 1995.

\bibitem[Neal \& Hinton(1998)Neal and Hinton]{neal1998view}
Neal, Radford~M and Hinton, Geoffrey~E.
\newblock A view of the em algorithm that justifies incremental, sparse, and
  other variants.
\newblock In \emph{Learning in graphical models}, pp.\  355--368. Springer,
  1998.

\bibitem[Neal et~al.(2011)]{neal2011mcmc}
Neal, Radford~M et~al.
\newblock Mcmc using hamiltonian dynamics.
\newblock \emph{Handbook of Markov Chain Monte Carlo}, 2\penalty0 (11), 2011.

\bibitem[Niculescu-Mizil \& Caruana(2005)Niculescu-Mizil and
  Caruana]{niculescu2005predicting}
Niculescu-Mizil, Alexandru and Caruana, Rich.
\newblock Predicting good probabilities with supervised learning.
\newblock In \emph{Proceedings of the 22nd international conference on Machine
  learning}, pp.\  625--632. ACM, 2005.

\bibitem[Opper \& Archambeau(2009)Opper and Archambeau]{opper2009variational}
Opper, Manfred and Archambeau, C{\'e}dric.
\newblock The variational gaussian approximation revisited.
\newblock \emph{Neural computation}, 21\penalty0 (3):\penalty0 786--792, 2009.

\bibitem[Peterson(1987)]{peterson1987mean}
Peterson, Carsten.
\newblock A mean field theory learning algorithm for neural networks.
\newblock \emph{Complex systems}, 1:\penalty0 995--1019, 1987.

\bibitem[Price(1958)]{price1958useful}
Price, Robert.
\newblock A useful theorem for nonlinear devices having gaussian inputs.
\newblock \emph{IRE Transactions on Information Theory}, 4\penalty0
  (2):\penalty0 69--72, 1958.

\bibitem[Rezende \& Mohamed(2015)Rezende and Mohamed]{rezende2015variational}
Rezende, Danilo~Jimenez and Mohamed, Shakir.
\newblock Variational inference with normalizing flows.
\newblock \emph{arXiv preprint arXiv:1505.05770}, 2015.

\bibitem[Rezende et~al.(2014)Rezende, Mohamed, and
  Wierstra]{rezende2014stochastic}
Rezende, Danilo~Jimenez, Mohamed, Shakir, and Wierstra, Daan.
\newblock Stochastic backpropagation and approximate inference in deep
  generative models.
\newblock \emph{arXiv preprint arXiv:1401.4082}, 2014.

\bibitem[Schulman et~al.(2015)Schulman, Levine, Abbeel, Jordan, and
  Moritz]{schulman2015trust}
Schulman, John, Levine, Sergey, Abbeel, Pieter, Jordan, Michael, and Moritz,
  Philipp.
\newblock Trust region policy optimization.
\newblock In \emph{Proceedings of the 32nd International Conference on Machine
  Learning (ICML-15)}, pp.\  1889--1897, 2015.

\bibitem[Settles(2010)]{settles2010active}
Settles, Burr.
\newblock Active learning literature survey.
\newblock \emph{University of Wisconsin, Madison}, 52\penalty0
  (55-66):\penalty0 11, 2010.

\bibitem[Simonyan \& Zisserman(2014)Simonyan and Zisserman]{simonyan2014very}
Simonyan, Karen and Zisserman, Andrew.
\newblock Very deep convolutional networks for large-scale image recognition.
\newblock \emph{arXiv preprint arXiv:1409.1556}, 2014.

\bibitem[Staines \& Barber(2012)Staines and Barber]{staines2012variational}
Staines, Joe and Barber, David.
\newblock Variational optimization.
\newblock \emph{arXiv preprint arXiv:1212.4507}, 2012.

\bibitem[Sun et~al.(2017)Sun, Chen, and Carin]{sun2017learning}
Sun, Shengyang, Chen, Changyou, and Carin, Lawrence.
\newblock Learning structured weight uncertainty in bayesian neural networks.
\newblock In \emph{Artificial Intelligence and Statistics}, pp.\  1283--1292,
  2017.

\bibitem[Williams(1992)]{williams1992simple}
Williams, Ronald~J.
\newblock Simple statistical gradient-following algorithms for connectionist
  reinforcement learning.
\newblock \emph{Machine learning}, 8\penalty0 (3-4):\penalty0 229--256, 1992.

\bibitem[Wu et~al.(2017)Wu, Mansimov, Liao, Grosse, and Ba]{wu2017scalable}
Wu, Yuhuai, Mansimov, Elman, Liao, Shun, Grosse, Roger, and Ba, Jimmy.
\newblock Scalable trust-region method for deep reinforcement learning using
  kronecker-factored approximation.
\newblock \emph{arXiv preprint arXiv:1708.05144}, 2017.

\end{thebibliography}
\bibliographystyle{icml2018}
\clearpage

\appendix
\section{Natural Gradient for Multivariate Gaussian}
\label{app:ng-gauss}
Suppose we have a model parameterized by $\params$ which lives in a subspace $\subspace$ (such as the set of symmetric matrices). The natural gradient $\natgrad_\params \loss$ is motivated in terms of a trust region optimization problem, that finding the optimal $\theta$ in a neighborhood of $\theta_0$ defined with KL divergence,
\begin{equation}
\label{eq:trust-region}
\begin{aligned}
&\phantom{=} \arg \min_{\params \in \subspace} \lrate (\nabla_\params \loss)^\transpose \params + \kldiv(p_\params \klbars p_\initParams) \nonumber \\
&\approx \arg \min_{\params \in \subspace} \lrate (\nabla_\params \loss)^\transpose \params + \frac{1}{2} (\params - \initParams)^\transpose \fisher (\params - \initParams)
\end{aligned}
\end{equation}
Then the optimal solution to this optimization problem is given by $\params -\lrate \fisher^{-1} \nabla_\params \loss$. Here $\fisher = \nabla^2_\params\, \kldiv(p_\params \klbars p_\initParams))$ is the Fisher matrix and $\lrate$ is the learning rate. Note that $\loss(\params)$ and $\kldiv(p_\params \klbars p_\initParams)$ are defined only for $\params, \initParams \in \subspace$, but these can be extended to the full space however we wish without changing the optimal solution.

Now let assume the model is parameterized by multivariate Gaussian $(\mean, \cov)$. The KL-divergence between $\normal(\mean, \cov)$ and $\normal_0(\mean_0, \cov_0)$ are:
\begin{equation}
\begin{aligned}
\kldiv(\normal \klbars \normal_{0}) &= \frac{1}{2} \left[ \log \frac{|\cov_0|}{|\cov|} - d + \tr (\cov_{0}^{-1} \cov) \right] \\
& + (\mean - \mean_0)^\transpose \cov_{0}^{-1} (\mean - \mean_0)
\end{aligned}
\end{equation}
Hence, the Fisher matrix w.r.t $\mean$ and $\cov$ are
\begin{equation}
\label{eq:fisher}
\begin{aligned}
\fisher_{\mean} &= \nabla_{\mean}^2\kldiv = \cov_{0}^{-1} \approx \cov^{-1} \\
\fisher_{\cov} &= \nabla_{\cov}^2\kldiv = \frac{1}{2}\cov^{-1} \otimes \cov^{-1}
\end{aligned}
\end{equation}
Then, by the property of vec-operator $(\mathbf{B}^\transpose \otimes \mathbf{A})\mathrm{vec}(\mathbf{X}) = \mathrm{vec}(\mathbf{A}\mathbf{X}\mathbf{B})$, we get the natural gradient updates
\begin{equation}
\begin{aligned}
\natgrad_{\mean}\loss &= \cov \nabla_{\mean} \loss \\
\natgrad_{\cov} \loss &= 2\cov \nabla_{\cov} \loss \cov
\end{aligned}
\end{equation}
An analogous derivation gives us $\natgrad_{\precision} \loss = 2\precision \nabla_{\precision} \loss \precision$. Considering $\cov = \precision^{-1}$, we have $\mathrm{d}\cov = -\cov \mathrm{d}\precision\cov$, which gives us the convenient formulas
\begin{equation}
\label{eq:ng-gauss}
\begin{aligned}
\natgrad_{\cov}\loss &= -2\nabla_{\precision}\loss \\
\natgrad_{\precision} \loss &= -2\nabla_{\cov}\loss
\end{aligned}
\end{equation}
Recall in variational inference, the gradient of ELBO $\elbo$ towards $\mean$ and $\cov$ are given as
 \begin{equation}
 \label{eq:sgd-gauss}
 \begin{aligned}
 \nabla_\mean \elbo %&= \nabla_\mean \expect[\log p(\data \given \weights)] + \klWeight \nabla_\mean \expect[\log p(\weights)] + \klWeight \nabla_\mean \entropy(q) \\
 &= \expect \left[ \nabla_\weights \log p(\data \given \weights) + \klWeight \nabla_\weights \log p(\weights) \right] \\
 \nabla_\cov \elbo %&= \nabla_\cov \expect[\log p(\data \given \weights)] + \klWeight \nabla_\cov \expect[\log p(\weights)] + \klWeight \nabla_\cov \entropy(q) \\
 &= \frac{1}{2} \expect \left[ \nabla^2_\weights \log p(\data \given \weights) + \klWeight \nabla^2_\weights \log p(\weights) \right] + \frac{\klWeight}{2} \cov^{-1}
 \end{aligned}
 \end{equation}
Based on eq.~\eqref{eq:sgd-gauss} and eq.~\eqref{eq:ng-gauss}, the natural gradient is given by:
\begin{equation}
\begin{aligned}
\natgrad_\mean \elbo &= \precision^{-1} \expect \left[ \nabla_\weights \log p(\data \given \weights) + \klWeight \nabla_\weights \log p(\weights) \right] \\
\natgrad_\precision \elbo &= -\expect \left[ \nabla^2_\weights \log p(\data \given \weights) + \klWeight \nabla^2_\weights \log p(\weights) \right] - \klWeight \precision
\end{aligned}
\end{equation}
% --------------------------------------------------------------------------------------------------- %
\section{Matrix Variate Gaussian}
\label{app:mvg}
Recently Matrix Variate Gaussian (MVG) distribution are also used in Bayesian neural networks \citep{louizos2016structured,sun2017learning}. A matrix variate Gaussian distributions models a Gaussian distribution for a matrix $\bm{W} \in \mathbb{R}^{n \times p}$,
\begin{equation}
\label{eq:mvg}
\begin{aligned}
	& p(\mathbf{W} | \mathbf{M}, \mathbf{U}, \mathbf{V}) \\ &= \frac{\exp (\frac{1}{2} \tr [\mathbf{V}^{-1} (\mathbf{W}-\mathbf{M})^\transpose  \mathbf{U}^{-1} (\mathbf{W}-\mathbf{M})])}{(2\pi)^{np/2} |\mathbf{V}|^{n/2} |\mathbf{U}|^{p/2}}
\end{aligned}
\end{equation}
In which $\mathbf{M} \in \mathbb{R}^{n \times p}$ is the mean, $\mathbf{U} \in \mathbb{R}^{n \times n}$ is the covariance matrix among rows and $\mathbf{V} \in \mathbb{R}^{p \times p}$ is the covariance matrix among columns. Both $\mathbf{U}$ and $\mathbf{V}$ are positive definite matrices to be a covariance matrix. Connected with Gaussian distribution, vectorization of $\mathbf{W}$ confines a multivariate Gaussian distribution whose covariance matrix is Kronecker product of $\mathbf{V}$ and $\mathbf{U}$.
\begin{equation}
\label{eq:mvg-vec}
\begin{aligned}
    \mathrm{vec}(\mathbf{W}) &\sim \mathcal{N} (\mathrm{vec}(\mathbf{M}), \mathbf{V} \otimes \mathbf{U})
\end{aligned}
\end{equation}

\section{Implementation Details}
\subsection{Regression Implementation Details}
The datasets were randomly splitted into training and test sets, with 90\% of the data for training and the remaining for testing. To reduce the randomness, we repeated the splitting process for 20 times (except two largest datasets, i.e., Year and Protein, where we repeated 5 times and 1 times, respectively.) For all datasets except two largest ones, we used neural networks with 50 hidden units. For two largest datasets, we used 100 hidden units. Besides, we also introduced a Gamma prior, $p(\tau) = Gam(a_0=6, b_0=6)$ for the precision of the Gaussian likelihood and included the posterior $q(\tau)=Gam(\alpha^{\tau}, \beta^{\tau})$ into variational objective. The variational posterior we used is $q(\weights, \lambda)=q(\weights)q(\tau), q(\tau) = \mathrm{Gamma}(\alpha^{\tau}, \beta^{\tau})$, then the expected likelihood $\mathcal{L}_r$ can be computed as 
\begin{equation}
\begin{aligned}
    \elbo_r &= \mathbb{E}_{q(\weights)}\mathbb{E}_{q(\tau)} \log p(y|\mathbf{x}, \weights, \tau) \\
    &= \mathbb{E}_{q(\weights)}\mathbb{E}_{q(\tau)} \log \mathcal{N}(y|\hat{y}(\mathbf{x}, \weights), \frac{1}{\tau}) \\
    &= \frac{1}{2}\mathbb{E}_{q(\weights)}\mathbb{E}_{q(\tau)} [\log \tau - \tau (y-\hat{y}(\mathbf{x}, \weights))^2 - \log 2\pi] \\
    &= \mathbb{E}_{q(\weights)} [\psi(\alpha^{\tau}) - \log \beta^{\tau} - \frac{\alpha^{\tau}}{\beta^{\tau}} (y - \hat{y}(\mathbf{x}, \weights))^2 - \log 2 \pi]
\end{aligned}
\end{equation}
Where $\psi$ represents digamma function. Therefore, ELBO can be computed with 
\begin{equation}
    \elbo = \elbo_r - \kldiv(q(\weights) \| p(\weights)) - \kldiv(q(\tau) \| p(\tau))
\end{equation}
With ELBO as above, we can directly compute the gradients towards variational parameters $\alpha, \beta$ using automatic differentiation.

In training, the input features and training targets were normalized to be zero mean and unit variance. We removed the normalization on the targets in test time. 
For each dataset, we set $\lrateMod = 0.01$ and $\lratePrecMod = 0.001$ unless state otherwise. We set batch size 10 for 5 small datasets with less than 2000 data points, 500 for Year and 100 for other fours. Besides, we decay the learning rate by $0.1$ in second half epochs.

\subsection{Classification Implementation Details}
Throughout classification experiments, we used VGG16 architecture but reduced the number of filters in each convolutional layer by half.

In training, we adopted learning rate selection strategy adopted by \citet{ba2016distributed}. In particular,  given a parameter update vector $\mathbf{v}$, the KL divergence between the predictive distributions before and after the update is given by the Fisher norm:
\begin{equation}
	\kldiv(q \klbars p) \approx \frac{1}{2} \mathbf{v}^{\top}\fisher\mathbf{v}
\end{equation}
Observe that choosing a step size of $\lrateMod$ will produce an update with squared Fisher norm $\lrateMod^2 \mathbf{v}^{\top}\fisher\mathbf{v}$. Motivated by the idea of trust region, we chose $\alpha$ in each iteration such that the squared Fisher norm is at most some value $c$:
\begin{equation}
	\lrateMod = \min \left(\lrateMod_{\text{max}}, \sqrt{\frac{c}{\mathbf{v}^{\top}\fisher\mathbf{v}}} \right)
\end{equation}
 We used an exponential decay schedule $c_k = c_0 \zeta^k$, where $c_0$ and $\zeta$ were tunable parameters ($c_0$ is 0.001 or 0.01 for noisy K-FAC in our CIFAR-10 experiments when models trained without/with Batch Normalization~\citep{ioffe2015batch}, $\zeta$ is 0.95; $c_0$ is 0.0001 for noisy Adam), and $k$ was incremented periodically (every epoch in our
CIFAR-10 experiments). In practice, computing $\mathbf{v}^{\top}\fisher\mathbf{v}$ involves curvature-vector products after each update which introduces significant computational overhead, so we instead used the approximate Fisher $\tilde{\fisher}$ that we used to compute natural gradient. The maximum step size $\lrateMod_{\text{max}}$ was set to be $0.01$.

To reduce computational overhead of K-FAC (also noisy K-FAC) introduced by updating approximate Fisher matrix $\tilde{\fisher}$ and inverting it, we set $T_{\text{stats}} = 10$ and $T_{\text{inv}} = 200$. That means our curvature statistics are somewhat more stale, but we found that it didn't significantly affect per-iteration optimization performance. $\lratePrecMod$ was set to 0.01 and 0.003 for noisy K-FAC and noisy Adam, respectively.

We noticed that it was favorable to tune regularization parameter $\klWeight$ and prior variance $\priorVar$. We used a small regularization parameter $\klWeight$ when data augmentation was adopted. E.g., we set $\klWeight = 0.1$ when models were trained with data augmentation while $\klWeight = 0.5$ otherwise. We speculate that using data augmentation leads to more training examples (larger $\ndata$), so it's reasonable to use a smaller $\klWeight$. Moreover, we set $\priorVar$ to $0.1$ when models were trained without Batch Normalization.

\subsection{Active Learning Implementation Details} 
Following the experimental protocol in PBP \citep{hernandez2015probabilistic}, we splited each dataset into training and test sets with 20 and 100 data points. All remaining data were included in pool sets.  In all experiments, we used a neural network with one hidden layer and 10 hidden units.

After fitting our model in training data, we evaluated the performance in test data and further added one data point from pool set into training set. The selection was based on the method described by which was equivalent to choose the one with highest predictive variance.
This process repeated 10 times, that is, we collected 9 points from pool sets. For each iteration, we re-trained the whole model from scratch.

Beyond that, as uncertainty estimation is of fundamental importance in active learning, we also performed experiments to evaluate the uncertainty estimation of our method directly, which was measured according to the Pearson's correlation of predictive variance compared to HMC~\citep{neal2011mcmc}. Recall Pearson's correlation,
\begin{equation}
    \rho(X, Y) = \frac{\mathrm{E}[(X-\mu_X)(Y-\mu_Y)]}{\sigma_X \sigma_Y}
\end{equation}
is a measure of linear correlation between two variables $X$ and $Y$. Pearson's corrleation ranges from $0$ to $1$, with bigger value representing stronger correlations. We compared several algorithms, including PBP, NNG-FFG, NNG-MVG, NNG-BlkTri.

We trained NNG-FFG, NNG-MVG and NNG-BlkTri for 20000 epochs, PBP for 40 epochs and HMC with 20 chains, 100000 iterations. For all the models, we used 1000 sampled weights for predicting on the testing set, thus we could compute the model's predicative variance for every data point in the test set. Finally, we computed the Pearson's correlation between different models and HMC in terms of predicative variance. In all experiments, we used $\lrateMod=0.01, \lratePrecMod=0.01$ and no extra damping term.

\subsection{Reinforcement Learning Implementation Details}
In all three tasks, CartPoleSwingup, MountainCar and DoublePendulum, we used one-layer Bayesian Neural Network with 32 hidden units for both BBB and NNG-MVG. And we used rectified linear unit (RELU) as our activation function. The number of samples drawn from variational posterior was fixed to 10 in the training process. For TRPO, the batch size was set to be 5000 and the replay pool has a fixed number of 100,000 samples. In both BBB and NNG-MVG, the dynamic model was updated in each epoch with 500 iterations and 10 batch size. For the policy network, one-layer network with 32 tanh units was used.

For all three tasks, we sparsified the rewards in the following way. A reward of $+1$ is given in CartPoleSwingup when $cos(\theta) > 0.8$, with $\theta$ the pole angle; when the car escapes the valley in MountainCar; and when $D < 0.1$, with $D$ the distance from the target in DoublePendulum.

To derive the intrinsic reward in \citet{houthooft2016vime}, we just need to analyze a single layer since we assume layer-wise independence in NNG-MVG. The intrinsic reward for each layer is given by (Note: $\elbo$ below is the ELBO with $q(\qParams)$ as prior.) 
\begin{equation}
\label{eq:intrinsic-reward}
\begin{aligned}
	& \kldiv \left(q(\qParams') \klbars q(\qParams)\right)  = \\
	& \frac{1}{2} \lrateMod^2
	\begin{bmatrix}
	\kvec \left\{\nabla_{\mean}\elbo\right\}\\\kvec \left\{\nabla_{\cov}\elbo\right\}
    \end{bmatrix}^T
    \begin{bmatrix}
	\fisher_{\mean}^{-1} &  \\  & \fisher_{\cov}^{-1}
	\end{bmatrix}
	\begin{bmatrix}
	\kvec \left\{\nabla_{\mean}\elbo\right\}\\\kvec \left\{\nabla_{\cov}\elbo\right\}
	\end{bmatrix}
\end{aligned}
\end{equation}
Where $\lrateMod$ is the step-size. As shown in eq.~\eqref{eq:fisher}, the Fisher matrix for $\mean$ is given by $\fisher_{\mean} = \cov^{-1}$, thus the first term in eq.~\eqref{eq:intrinsic-reward} is easy to get by exploiting Kronecker structure. However, $\fisher_{\cov} = \frac{1}{2}\cov^{-1} \otimes \cov^{-1}$ where $\cov$ itself is a gigantic matrix which makes computation of the second term intractable. Fortunately, the approximate variational posterior is a matrix variate Gaussian whose covariance is a Kronecker product, i.e.~$\matrixNormal(\mathbf{W} ; \mathbf{M}, \cov_1, \cov_2) = \normal(\kvec(\mathbf{W}); \kvec(\mathbf{M}), \cov_2 \otimes \cov_1)$, where $\mathbf{W}$ is of size $m \times n$ and $\mean=\kvec(\mathbf{M})$. 

Using $\nabla_{\cov}\elbo = \nabla_{\cov_2}\elbo \otimes \cov_1 + \cov_2 \otimes \nabla_{\cov_1}\elbo$ and substitute $\fisher_{\cov}$ with $\frac{1}{2} \cov^{-1} \otimes \cov^{-1}$, we get the following identity
\begin{equation}
\label{eq:identity}
\begin{aligned}
	& \kvec \left\{\nabla_{\cov}\elbo \right\}^T \fisher_{\cov} \kvec \left\{\nabla_{\cov}\elbo \right\} = \\
	& \begin{bmatrix}
	\kvec \left\{\nabla_{\cov_1}\elbo\right\}\\\kvec \left\{\nabla_{\cov_2}\elbo\right\}
    \end{bmatrix}^T
    \begin{bmatrix}
	\fisher_{\cov_1} & \fisher \\ \fisher^T & \fisher_{\cov_2}
	\end{bmatrix}
	\begin{bmatrix}
	\kvec \left\{\nabla_{\cov_1}\elbo\right\}\\\kvec \left\{\nabla_{\cov_2}\elbo\right\}
	\end{bmatrix}
\end{aligned}
\end{equation}
where Fisher matrices $\fisher_{\cov_1}$ and $\fisher_{\cov_2}$
\begin{equation}
\begin{aligned}
	\fisher_{\cov_1} &= \frac{n}{2} \left(\cov_1^{-1} \otimes \cov_1^{-1}\right) \\
	\fisher_{\cov_2} &= \frac{m}{2} \left(\cov_2^{-1} \otimes \cov_2^{-1}\right) \\
\end{aligned}
\end{equation}
By further ignoring off-diagonal block $\fisher$ in eq.~\eqref{eq:identity}, we can decompose $\kvec \left\{\nabla_{\cov}\elbo \right\}^T \fisher_{\cov}^{-1} \kvec \left\{\nabla_{\cov}\elbo \right\}$ into two terms,
\begin{equation}
\begin{aligned}
	& \kvec \left\{\nabla_{\cov_1}\elbo \right\}^T \fisher_{\cov_1}^{-1} \kvec \left\{\nabla_{\cov_1}\elbo \right\} \\ & = \frac{2}{n}\kvec \left\{\nabla_{\cov_1}\elbo \right\}^T \kvec \left( \cov_1^{-1}\nabla_{\cov_1}\elbo\cov_1^{-1} \right)
\end{aligned}
\end{equation}
and
\begin{equation}
\begin{aligned}
	& \kvec \left\{\nabla_{\cov_2}\elbo \right\}^T \fisher_{\cov_2}^{-1} \kvec \left\{\nabla_{\cov_2}\elbo \right\} \\ & = \frac{2}{m}\kvec \left\{\nabla_{\cov_2}\elbo \right\}^T \kvec \left( \cov_2^{-1}\nabla_{\cov_2}\elbo\cov_2^{-1} \right)
\end{aligned}
\end{equation}
Now, each term can be computed efficient since $\cov_1$ and $\cov_2$ are small matrices.

\section{Additional Results}
\begin{figure}[h]
  \centering
  {\includegraphics[width=0.9\linewidth]{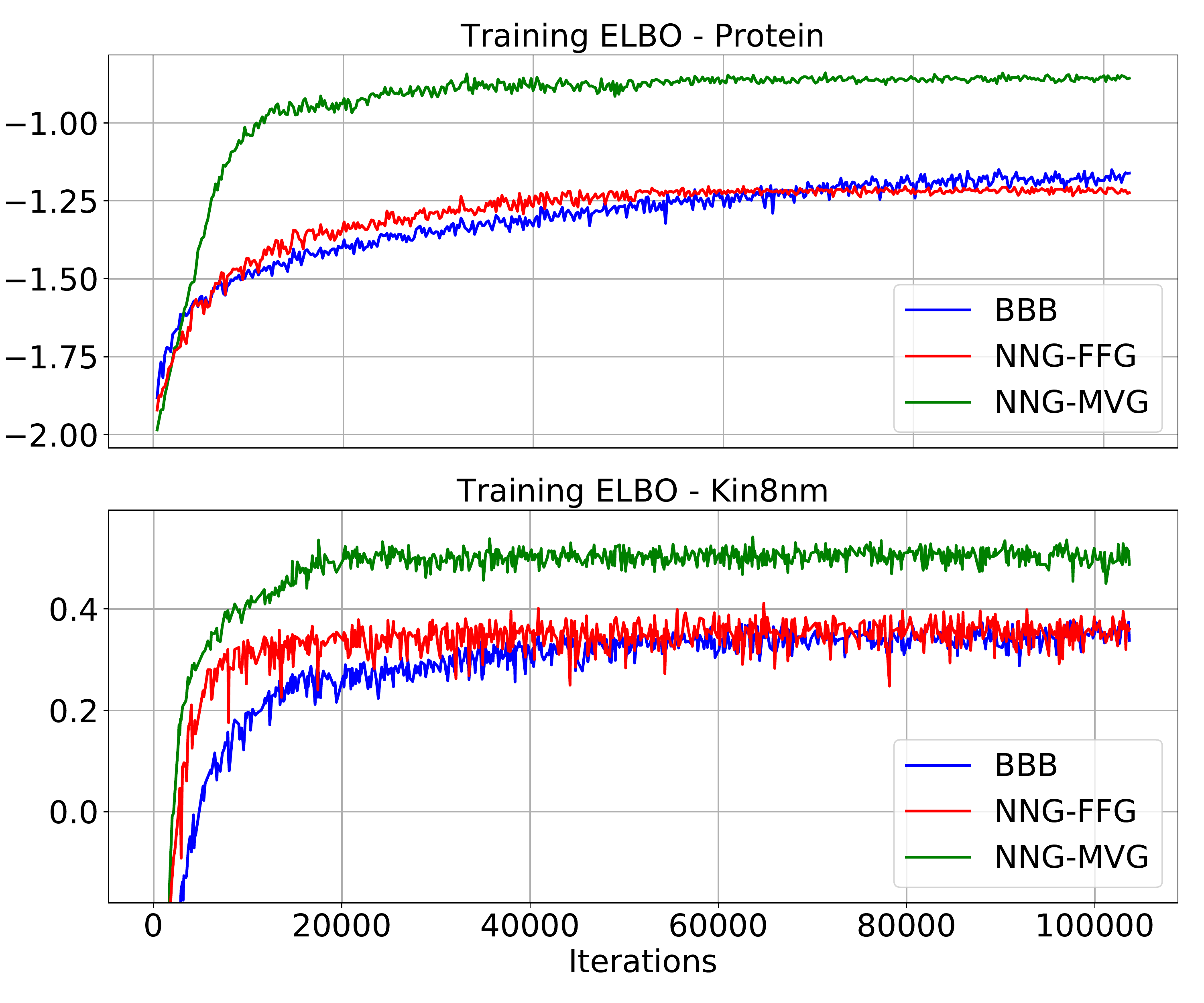}}
  \caption{Training curves for all three methods. For each method, we tuned the learning rate for updating the posterior mean. Note that BBB and NNG-FFG use the same form of $q$, while NNG-MVG uses a more flexible $q$ distribution.
\label{fig:convergence}}
\end{figure}
We also run PBP\_MV~\citep{sun2017learning} and VMG~\citep{louizos2016structured} on regression datasets from UCI collection~\citep{asuncion2007uci}. Results are shown in Table~\ref{tab:uci}. Note that VMG introduced pseudo input-output pairs to enhance the flexibility of posterior distribution.
\begin{table}[h]
\caption{Averaged test RMSE and log-likelihood for the regression benchmarks.}
\vspace{-1em}
\label{tab:uci}
\begin{center}
\resizebox{\columnwidth}{!}{
\begin{sc}
\begin{tabular}{lcccc}
\toprule
\textbf{}        & \multicolumn{2}{c}{Test RMSE} & \multicolumn{2}{c}{Test log-likelihood} \\
\textbf{Dataset} & PBP\_MV & VMG & PBP\_MV & VMG \\
\midrule
Boston  		 & 3.137$\pm$0.155 & 2.810$\pm$0.110 & -2.666$\pm$0.081 & -2.540$\pm$0.080 \\ 
Concrete         & 5.397$\pm$0.130 & 4.700$\pm$0.140 & -3.059$\pm$0.029 & -2.980$\pm$0.030 \\  
Energy           & 0.556$\pm$0.016 & 1.160$\pm$0.030 & -1.151$\pm$0.016 & -1.450$\pm$0.030 \\ 
Kin8nm           & 0.088$\pm$0.001 & 0.080$\pm$0.001 &  1.053$\pm$0.012 &  1.140$\pm$0.010 \\
Naval            & 0.002$\pm$0.000 & 0.000$\pm$0.000 &  4.935$\pm$0.051 &  5.840$\pm$0.000 \\
Pow. Plant       & 4.030$\pm$0.036 & 3.880$\pm$0.030 & -2.830$\pm$0.008 & -2.780$\pm$0.010 \\
Protein          & 4.490$\pm$0.012 & 4.140$\pm$0.010 & -2.917$\pm$0.003 & -2.840$\pm$0.000 \\
Wine             & 0.641$\pm$0.006 & 0.610$\pm$0.010 & -0.969$\pm$0.013 & -0.930$\pm$0.020 \\
Yacht            & 0.676$\pm$0.054 & 0.770$\pm$0.060 & -1.024$\pm$0.025 & -1.290$\pm$0.020 \\
Year             & 9.450$\pm$NA & 8.780$\pm$NA & -3.392$\pm$NA & -3.589 $\pm$NA \\ 
\bottomrule
\end{tabular}
\end{sc}
}
\end{center}
\end{table}

While optimization was not the primary focus of this work, we compared NNG with the baseline BBB~\citep{blundell2015weight} in terms of convergence. 
Training curves for two regression datasets are shown in Figure~\ref{fig:convergence} . 
We found that NNG-FFG trained in fewer iterations than BBB, while leveling off to similar ELBO values, even though our BBB implementation used Adam, and hence itself exploited diagonal curvature.
Furthermore, despite the increased flexibility and larger number of parameters, NNG-MVG took roughly 2 times fewer iterations to converge, while at the same time surpassing BBB by a significant margin in terms of the ELBO.

\end{document}